\pdfoutput=1
\documentclass[a4paper,fleqn]{cas-sc}

\usepackage[utf8]{inputenc}
\usepackage{kotex}
\usepackage[english]{babel}
\usepackage{longtable,booktabs,array,calc,ragged2e,graphicx,xurl}
\providecommand{\tightlist}{\setlength{\itemsep}{0pt}\setlength{\parskip}{0pt}}

\hypersetup{pdftitle={MARO: Multi-Agent Retrieval-Augmented Question Answering over Korean Policy Reports and Its Protocol-Sensitive Evaluation},pdfauthor={Saksonita Khoeurn, In Hwan Oh, Sovandara Chhoun, Phannet Pov, Wan-Sup Cho},pdfkeywords={Retrieval-augmented generation, Evaluation methodology, LLM-as-a-judge, Abstention, Multi-agent orchestration, Korean question answering, Knowledge graph}}

\ExplSyntaxOn
\RenewDocumentEnvironment{Abstract}{o}
{
 \group_begin:
 \IfNoValueTF{#1}{}{\tex_gdef:D\abstractname{#1}}
 \parindent=0pt
 \box_if_empty:NTF\g_stm_key_box
 {\leftskip=.35\textwidth}
 {
  \dim_gset:Nn\l_tmpa_dim{\box_ht:N\g_stm_key_box}
  \dim_gadd:Nn\l_tmpa_dim{\box_dp:N\g_stm_key_box}
  \leftskip=.35\textwidth
  \hspace*{-.35\textwidth}
  \noindent\hbox_to_wd:nn{0pt}{\box\g_stm_key_box\hss}
  \skip_vertical:n{-\l_tmpa_dim}
 }
 \noindent\abstractname\par
 \skip_vertical:n{-4pt}
 \noindent\rule{.65\textwidth}{.2pt}\par\footnotesize
 \ignorespaces\everypar{\parindent=1.5em}
}{\par\group_end:}
\cs_set:Npn \__first_footerline:
{
  \group_begin:
  \small
  \sffamily
  \__short_authors: :~
  { \rmfamily \itshape Preprint~ submitted ~to ~Array }
  \group_end:
}
\ExplSyntaxOff

\begin{document}
\let\WriteBookmarks\relax
\shorttitle{MARO: Protocol-sensitive evaluation of multi-agent RAG}
\shortauthors{S. Khoeurn et al.}
\title[mode=title]{MARO: Multi-Agent Retrieval-Augmented Question Answering over Korean Policy Reports and Its Protocol-Sensitive Evaluation}

\author[1,2]{Saksonita Khoeurn}[orcid=0000-0002-8655-5896]
\cormark[1]
\ead{saksonita@chungbuk.ac.kr}
\author[3]{In Hwan Oh}
\author[3]{Sovandara Chhoun}
\author[3]{Phannet Pov}
\author[1,2]{Wan-Sup Cho}[orcid=0000-0002-4395-1979]

\affiliation[1]{organization={Department of Management Information Systems, Chungbuk National University},
  city={Cheongju}, country={Republic of Korea}}
\affiliation[2]{organization={BigDataLabs}, city={Cheongju}, country={Republic of Korea}}
\affiliation[3]{organization={Department of Big Data, Chungbuk National University},
  city={Cheongju}, country={Republic of Korea}}
\cortext[1]{Corresponding author}

\begin{abstract}
Policy researchers need answers from long Korean reports that point to the passages behind them. We present MARO, a multi-agent retrieval-augmented generation (RAG) system that answers Korean policy questions from 245 regional policy-research reports. Five language-model agents plan the question, retrieve passages through dense search and an ontology-guided knowledge graph, select among candidate drafts, write a cited answer and check its claims, and every stage's record is saved with the answer. We evaluate MARO against vanilla RAG, LightRAG and simplified variants on 100 expert-reviewed questions, 10 of them unanswerable, using repeated generations, two generator families, two judge packets, blinded human ratings and three-way refusal coding.

MARO's answers are more complete. On answerable questions, answer-only grading favors it over vanilla RAG by +0.226 on a 0--2 scale (95\% interval +0.115 to +0.344), and blinded human ratings of one run agree and also favor its reporting periods. Its correctness advantage is not robust: the paired contrast is +0.060 (\ensuremath{-}0.080 to +0.200) under trace-aware grading, \ensuremath{-}0.010 under answer-only grading and \ensuremath{-}0.118 with the second generator. Two simplified paths without planning, selection or verification gain a similar completeness advantage, while MARO uses at least 3.4 times the model calls. MARO hedges on unanswerable questions; vanilla RAG refuses them but also refuses about a fifth of answerable ones.

MARO's value therefore lies in fuller, better-dated and checkable answers rather than higher correctness. Such systems need repeated, paired, criterion-specific evaluation with explicit refusal coding.
\end{abstract}
\maketitle

\section{Introduction}

Policy researchers who work with long regional reports need answers they can check: a figure for the right indicator, year and region, together with the passage it comes from. Retrieval-augmented generation (RAG) answers such questions by combining language generation with evidence retrieved from an external collection, and agentic and multi-agent RAG systems go further, decomposing a question, retrieving through several paths, combining candidate answers and verifying the result (Nguyen et al., 2025). We built MARO (Multi-Agent Retrieval Orchestrator), a five-agent RAG system that answers Korean policy questions from 245 regional policy-research reports and cites a numbered source passage for every claim.

Whether such a system answers better than simpler RAG is harder to establish than it looks. An assessment may consider whether relevant evidence was retrieved, whether claims are supported, whether the answer is complete, and whether the system declines an unanswerable question. Frameworks such as RAGAS, ARES and RAGChecker address several of these dimensions (Es et al., 2024; Saad-Falcon et al., 2024; Ru et al., 2024). Even with multiple metrics, a comparison between systems can depend on the generation run, judge, scoring population and retrieval units used, and multi-stage systems add cost and more opportunities for variation. A useful evaluation must therefore establish not only whether one configuration obtains a higher score, but also whether the difference persists across reasonable evaluation choices and which aspects of answer quality account for it.

Recent work has begun to examine the robustness of RAG comparisons. Zeng et al.~(2025) investigate how biases in LLM-based judging affect reported GraphRAG gains. Akarsu et al.~(2026) vary the embedder, corpus, and judge in a study of RAG pipelines and show that conclusions about citation performance depend on where citations are measured. Ferrazzi et al.~(2026) compare enhanced and agentic RAG across scenarios and report trade-offs between effectiveness and cost. Together, these studies show that a single ranking may not adequately characterize a system comparison. They also motivate closer examination of additional sources of variation: repeated generation, the material presented to a judge, the treatment of unanswerable questions, and the units used to measure retrieval.

This paper presents MARO and evaluates it in that spirit. Its role-specialized design follows established agentic RAG patterns and is not claimed as a new architecture. What we contribute is a documented, traceable application to Korean policy reports and an evaluation that asks which of its advantages survive changes in the evaluation protocol. We distinguish changes in observed point estimates from evidence of a reliable difference between systems or protocols.

\subsection{Application setting}

Korean regional policy reports provide a demanding setting for evidence-grounded question answering. Questions may require a system to identify a specific indicator, reporting period, and administrative region across lengthy documents and tables. A passage can be topically relevant while reporting a value for the wrong year or location. This setting reflects the importance of temporal and geographic context in question answering (Zhang \& Choi, 2021) and the need to assess whether generated claims are supported by appropriate evidence (Gao et al., 2023).

The system must also distinguish questions answerable from the document collection from those for which the collection does not provide sufficient evidence. Abstention is not always a simple binary outcome: a response may acknowledge missing evidence while nevertheless supplying adjacent or speculative information. Prior work has examined unanswerability in RAG and partial abstention in language models (Peng et al., 2025; Wen et al., 2025). We apply a three-way coding scheme---clean refusal, hedge, and answer---to make these behaviors visible in the system comparison.

\subsection{MARO and research questions}

MARO is a five-stage pipeline: a planner, a retriever with candidate drafting, a selector, a synthesizer and a verifier (Section 3). Retrieval combines dense search with a knowledge graph of 114,304 entities extracted from the same reports. Each stage's input and output are saved with the answer, so every answer can be inspected stage by stage. We compare MARO with vanilla RAG, LightRAG and simplified variants of itself on 100 expert-reviewed questions over Korean regional policy reports.

The study addresses five questions:

\begin{enumerate}
\def\labelenumi{\arabic{enumi}.}
\tightlist
\item
  \textbf{Answer quality and cost:} On which criteria do MARO's answers differ from those of simpler RAG systems, and at what cost in model calls?
\item
  \textbf{Repetition and generator:} How do observed system contrasts vary across generation runs and two generator families?
\item
  \textbf{Judge input and scoring population:} How do results differ when the judge receives different information, and when correctness is summarized over answerable questions or the full question set?
\item
  \textbf{Refusal behavior:} What distinctions are exposed by coding clean refusals, hedges, and answers separately?
\item
  \textbf{Retrieval measurement:} How do comparisons of evidence retrieval differ when results are counted by source unit or measured at matched source-text budgets?
\end{enumerate}

The principal pipeline and vanilla systems differ in retrieval, prompting, and orchestration. Their comparison therefore estimates differences between complete configurations; it does not isolate the causal effect of adding agents.

\subsection{Contributions and scope}

This paper makes five contributions.

First, it describes MARO, a multi-agent RAG system for Korean policy reports in which every claim cites a numbered passage and every stage's record is saved, and traces one multi-hop question through it (Section 3).

Second, it reports a paired, repeated comparison showing how correctness estimates for the studied systems vary across generation and evaluation settings. Several point estimates change in magnitude or sign, while their uncertainty intervals include zero. These results caution against treating an observed ranking from a single protocol as conclusive.

Third, it examines judge-packet sensitivity using identical final answers. Removing a diagnostic verifier trace from the saved record changes some grades despite leaving the answers unchanged. A separate answer-only regrading of the principal comparison provides an additional protocol check; because that audit also changes source visibility and the serving endpoint, it is not a causal estimate of trace influence alone.

Fourth, it separates refusal, hedge, and answer behavior and relates these categories to both answerable and unanswerable questions. This analysis reveals different error profiles: in the tested runs, the pipeline often hedges on unanswerable questions, while vanilla RAG also refuses some answerable questions.

Fifth, it introduces an expert-reviewed benchmark of 100 questions over 245 published Korean policy reports rather than synthetic documents, including 214 gold-evidence annotations and documented searches for the 10 unanswerable items. The benchmark and saved evaluation artifacts provide a basis for examining the reported comparisons and their limitations.

The study's evidence is bounded by one corpus, a small unanswerable subset, the selected systems and models, and a human evaluation of one generation run. Within those boundaries, the results show that completeness, correctness, refusal behavior, and cost can lead to different interpretations of the same comparison. We therefore recommend reporting them separately alongside uncertainty and the details of the evaluation protocol.

Section 2 reviews related work. Section 3 describes MARO, Section 4 the benchmark and the compared systems, and Section 5 the evaluation protocol. Section 6 shows how each evaluation choice changes the result, and Section 7 reports the human study. Section 8 discusses implications and limitations, and Section 9 concludes.

\section{Related work}

Prior work shows that RAG comparisons can depend on judging, retrieval budgets and cost accounting. It has not combined generation repetition, generator replication, judge-packet control and three-way abstention coding in one comparison. This section places each of our evaluation choices against that work.

\subsection{Audits of RAG architecture comparisons}

Retrieval-augmented generation combines a language model with retrieved text (Lewis et al., 2020); graph variants organize information across passages (Edge et al., 2024; Guo et al., 2025). Several studies question reported gains of these variants. Han et al.~(2025) compared RAG and GraphRAG under a unified protocol and found that matching retrieved-token budgets closes most of the gap. Zeng et al.~(2025) showed that position, length and trial biases in pairwise judging inflate GraphRAG win rates over naive RAG. GraphRAG-Bench (Xiang et al., 2026) and Fan et al.~(2026) report that graph retrieval helps mainly on complex or multi-hop questions.

The closest study to ours is Akarsu et al.~(2026). Holding five pipelines fixed, they varied the embedder, corpus and judge, and found that changing where citations are measured inverts the architecture ranking. They argue that robustness on all three axes is a minimum bar for architecture claims. Their pipelines share one generator and synthesis prompt and differ only in retrieval: vanilla dense search, a typed-edge walk over links between corpus items, and an iterative agentic search loop. MARO instead plans once and runs a fixed sequence of parallel retrieval, candidate selection, synthesis and verification over an entity graph extracted from the text, so our principal contrast compares whole systems rather than retrievers alone. We extend this line in four ways. We vary generation runs and generator family rather than the embedder. We control the judge packet while holding answer strings fixed. And we code abstention in three categories and validate with blinded human ratings. Finally, their main corpus is synthetic, generated by the same model family that answers and judges, whereas ours consists of 245 published policy reports, with every question, reference answer and evidence passage checked by a domain expert. In policy QA specifically, Mathur et al.~(2026) found that better retrieval did not reliably produce better answers.

Our protocol-sensitivity profile (Section 5.6) follows multiverse and specification-curve analysis, which report a result across every defensible combination of analytic choices instead of one chosen path (Steegen et al., 2016; Simonsohn et al., 2020). Bell et al.~(2022) brought multiverse analysis to machine-learning claims, where the choices are hyperparameters. In our setting the choices are evaluation decisions: the generation run, generator, judge packet, scoring population and criterion. A recent survey of RAG evaluation organizes the metrics, datasets and frameworks used for retrieval, generation and system-level assessment (Gan et al., 2025); the profile asks instead how far one comparison depends on which of these choices is made.

\subsection{Multi-stage and agentic RAG versus simpler baselines}

MA-RAG (Nguyen et al., 2025) assigns planning, step definition, extraction and answering to specialized agents. MAIN-RAG (Chang et al., 2025) uses multiple agents to filter retrieved documents. Self-RAG (Asai et al., 2024) and Corrective RAG (Yan et al., 2024) add critique or corrective retrieval. These papers report gains from single evaluation runs.

A recent survey classifies agentic RAG into single-agent, multi-agent and hierarchical designs (Singh et al., 2025). Graph-based retrieval also continues to develop: HippoRAG 2 extends personalized PageRank over a knowledge graph with deeper passage integration and online use of an LLM (Guti\'errez et al., 2025). Our external graph baseline is LightRAG (Section 4.7).

Cost-aware comparisons are less favorable. Kapoor et al.~(2025) argued that agent evaluations must report cost alongside accuracy, and showed that simple baselines can match complex agents. Ferrazzi et al.~(2026) found neither enhanced nor agentic RAG universally superior, with agentic RAG up to 3.6 times more expensive. Dimitrova (2025) reported that multi-agent configurations underperformed single-agent baselines on an agricultural QA set. We likewise report model calls and tokens alongside accuracy, and examine whether the accuracy side of the trade-off depends on the evaluation protocol.

\subsection{Run-to-run variance and statistical practice}

LLM outputs vary across runs even under settings intended to be deterministic, and this variation can change model rankings (Atil et al., 2025). Madaan et al.~(2024) quantified seed variance in common benchmarks. Hochlehnert et al.~(2025) showed that on small benchmarks, reported improvements can fall within run-to-run variation. Heineman et al.~(2025) separate a benchmark's signal, its ability to tell stronger models from weaker ones, from its noise, its sensitivity to random variation between training checkpoints, and show that benchmarks with a higher signal-to-noise ratio support more reliable small-scale decisions. Miller (2024) recommended question-level paired differences and clustered standard errors. Our estimator follows these recommendations. We additionally generate each configuration three times to test whether architecture contrasts are stable across runs.

\subsection{LLM-as-a-judge reliability and judge input}

LLM judges show position, verbosity and self-enhancement biases (Zheng et al., 2023). Ye et al.~(2025) catalogue twelve such biases and quantify each by systematically modifying the judged content. Marioriyad et al.~(2026) showed that metadata cues injected into a judge prompt shift verdicts, usually without the judge acknowledging them. Agent-as-a-Judge (Zhuge et al., 2025) argues that letting a judge inspect intermediate steps improves agreement with humans. Trace-aware and trace-blind judges can also disagree sharply about whether an answer is grounded (Elkoussy \& Perez, 2026). Judge verdicts can also vary across repeated calls at temperature 0 (Tamba, 2026).

We found no study that holds a final answer fixed and varies only whether the judge sees the pipeline trace. Our C/E comparison does this by construction, and our answer-only audit tests the principal comparison under a common packet.

\subsection{Abstention and refusal in RAG}

The Retrieval-Augmented Generation Benchmark (RGB) introduced negative rejection, testing whether a RAG system declines when retrieved documents lack the answer (Chen, Lin, et al., 2024). UAEval4RAG scores answerable and unanswerable requests jointly and shows that RAG components shift the trade-off between them (Peng et al., 2025). RefusalBench found that grounded models are often either overconfident or overcautious (Muhamed et al., 2026). Search augmentation can also reduce abstention on unanswerable questions (Xie et al., 2026).

Most of these benchmarks treat abstention as binary. The abstention survey by Wen et al.~(2025) defines partial abstention, where a response declines and then answers anyway, and R\"ottger et al.~(2024) annotated partial refusal. Our hedge category applies this concept to architecture comparison.

\subsection{Comparing retrieval across different source units}

Recall@k assumes that systems return comparable units. Graph-based systems return entities, summaries or expanded chunks of different lengths. Han et al.~(2025) and NodeRAG (Xu et al., 2025) therefore compare systems at matched token budgets, and HiChunk (Lu et al., 2026) reports evidence recall across budgets. We follow this logic with gold-evidence recall at matched source-text character budgets (Section 5.5).

\subsection{Korean and public-sector benchmarks}

Korean RAG resources include SDS KoPub VDR for visual retrieval over public documents (Lee et al., 2025), Ko-LongRAG for long-context RAG with unanswerable items (Kim et al., 2025), the Allganize Korean RAG evaluation dataset (Allganize, 2024), and KBL, a benchmark for Korean legal language understanding (Kim et al., 2024). Outside Korea, PLLuM's RAG instruction-following set includes refusal conditions over Polish public-administration documents (Koco\'n et al., 2025). We found no Korean regional policy-report benchmark with typed questions, gold evidence and unanswerable items. Multilingual retrieval in our systems uses BGE-M3 embeddings (Chen, Xiao, et al., 2024).

\section{The MARO system}

MARO answers a Korean policy question from the report corpus in five stages and saves each stage's record with the answer. This section describes the pipeline (Section 3.1), the indexes and the design of each stage (Section 3.2), one question traced end to end (Section 3.3), and the model calls and generation settings (Section 3.4).

\subsection{Pipeline overview}

MARO separates orchestration from the retrieval backend and records each stage's output (Figure 1).

\begin{enumerate}
\def\labelenumi{\arabic{enumi}.}
\tightlist
\item
  \textbf{Planner.} Assigns retrieval modes and subqueries, combining fast rules with model calls when needed. The adapter deduplicates query--mode tasks and allows at most four retrieval calls.
\item
  \textbf{Retriever with candidate drafting.} Each task calls the backend and returns passages, source identifiers, graph context where applicable, and a candidate answer. In the local path, vector search and reranking select passages.
\item
  \textbf{Selector.} Evaluates candidate responses and chooses a selection or combination strategy; a single candidate bypasses the model-based comparison.
\item
  \textbf{Synthesizer.} Produces the final cited answer from candidates and merged sources.
\item
  \textbf{Verifier.} Checks the answer against retrieved material. In C its verdict is diagnostic and never changes the answer. In G and I, a failed check can replace the answer with a refusal.
\end{enumerate}

\begin{figure}
\centering
\includegraphics[width=1\linewidth,height=\textheight,keepaspectratio]{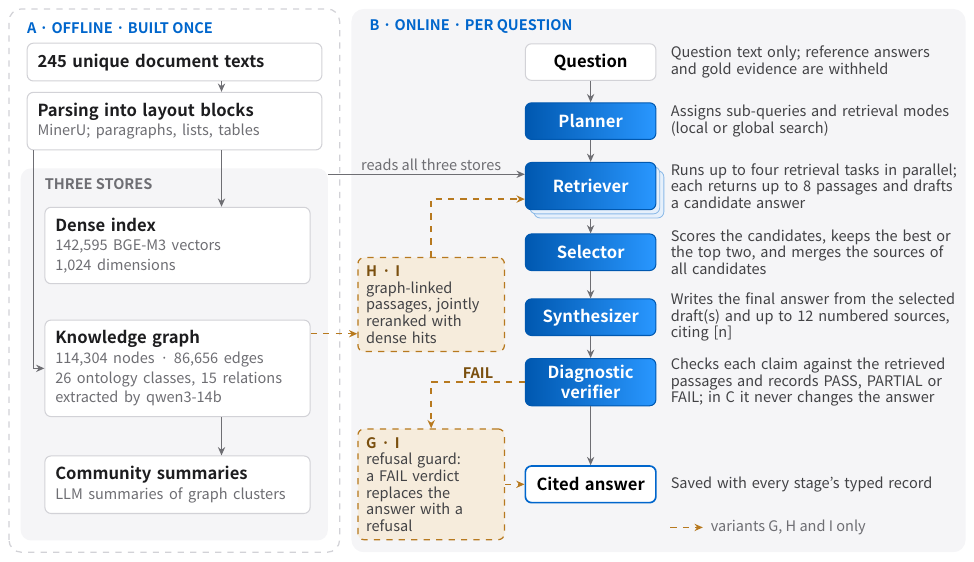}
\caption{MARO architecture. (A) Offline corpus preparation builds the dense index, the knowledge graph and community summaries from 245 documents. (B) Online, each question passes through the planner, retriever, selector, synthesizer and diagnostic verifier. G and I add an optional refusal guard; H and I jointly rerank graph-linked passages with dense hits.}
\end{figure}

\subsection{Indexing and stage design}

\textbf{Offline indexing.} MARO answers from three stores built once from the parsed corpus (Section 4.1) and frozen before any question was run. The dense index embeds every layout block with BGE-M3. The knowledge graph is extracted block by block: qwen3-14b reads each block with a domain ontology of 26 entity classes (for example Organization, Policy, Project, Region and Metric) and 15 typed relations (for example FUNDS, IMPLEMENTS, TARGETS and MEASURES). A validator discards triples that violate a relation's domain or range, and a normalizer merges aliases of the same entity. Every node and edge keeps the identifiers of the blocks it came from, so graph results can always be traced to original passages. Communities are found by modularity-based clustering of the graph, summarized by the LLM, and the summaries are embedded for global search.

\textbf{Stage interfaces.} Each stage passes a typed record to the next, and every record is saved with the answer. Table 1 lists what each stage reads, what it writes and when it calls the model.

\begin{longtable}[]{@{}
  >{\raggedright\arraybackslash}p{(\linewidth - 6\tabcolsep) * \real{0.1250}}
  >{\raggedright\arraybackslash}p{(\linewidth - 6\tabcolsep) * \real{0.2500}}
  >{\raggedright\arraybackslash}p{(\linewidth - 6\tabcolsep) * \real{0.3125}}
  >{\raggedright\arraybackslash}p{(\linewidth - 6\tabcolsep) * \real{0.3125}}@{}}
\caption{Inputs, outputs and model calls of the five online stages in configuration C.}\tabularnewline
\toprule\noalign{}
\begin{minipage}[b]{\linewidth}\raggedright
Stage
\end{minipage} & \begin{minipage}[b]{\linewidth}\raggedright
Input
\end{minipage} & \begin{minipage}[b]{\linewidth}\raggedright
Output
\end{minipage} & \begin{minipage}[b]{\linewidth}\raggedright
Model call (temperature)
\end{minipage} \\
\midrule\noalign{}
\endfirsthead
\toprule\noalign{}
\begin{minipage}[b]{\linewidth}\raggedright
Stage
\end{minipage} & \begin{minipage}[b]{\linewidth}\raggedright
Input
\end{minipage} & \begin{minipage}[b]{\linewidth}\raggedright
Output
\end{minipage} & \begin{minipage}[b]{\linewidth}\raggedright
Model call (temperature)
\end{minipage} \\
\midrule\noalign{}
\endhead
\bottomrule\noalign{}
\endlastfoot
Planner & Question & Intent (factoid, thematic, comparative or chart); one or two sub-queries; retrieval modes & Only when keyword rules do not decide: intent classification (0.0); decomposition of comparative questions (0.1) \\
Retriever & (sub-query, mode) tasks, at most four, run in parallel & Per task: up to 8 passages with identifiers, graph context, one candidate answer & One drafting call per task, issued by the retrieval backend (0.1) \\
Selector & Candidate answers & A 0--10 score and reason per candidate; strategy select\_best or synthesize & Only with two or more candidates (0.0) \\
Synthesizer & Question, selected draft(s), up to 12 merged and numbered sources & Final answer with {[}n{]} citations & One call (0.1) \\
Verifier & Final answer, original passages of all candidates & Per-claim support, faithfulness score, PASS, PARTIAL or FAIL & One call (0.0) \\
\end{longtable}

\textbf{Planning.} The planner first applies Korean keyword rules. Words such as 비교 (compare), 차이 (difference) or 대비 (contrast) mark a comparative question; 전반적 (overall), 정책 방향 (policy direction) or 시사점 (implications) mark a thematic one. Only when no rule fires does an LLM call classify the intent. Factoid questions are kept whole and sent to local search. Thematic questions go to both global and local search. Comparative questions are split into two sub-queries, by a rule for the pattern ``A와 B를 비교'' (compare A and B) and otherwise by an LLM call. The retrieval adapter maps the legacy hybrid mode to local, removes duplicate tasks and runs at most four in parallel.

\textbf{Local and global search.} Local search over-fetches 24 blocks by cosine similarity to the sub-query and keeps the top 8 after cross-encoder reranking. Graph entities extracted from those 8 blocks, together with entities whose names match query tokens, seed a traversal restricted to relation types suited to the question; a question about funding, for example, follows FUNDS and CONDUCTS. The resulting triples enter the context as graph context but do not compete with passages for rank. Global search ranks community summaries against the query, reranks them, and returns the reranked blocks behind the top five communities. For each task the backend then drafts a candidate answer from its context in one LLM call.

\textbf{Selection and source merging.} With one candidate the selector passes it through without a model call. With several, one LLM call scores each candidate from 0 to 10 and gives a one-sentence reason. If the best score reaches 7.5, that candidate alone is selected (select\_best); otherwise the top two are kept for merging (synthesize). Either way, the sources of all candidates are merged, the selected candidate's first. Duplicates are removed by chunk identifier and by passage text, because several reports were ingested twice, and up to 12 sources are numbered {[}1{]}--{[}12{]}.

\textbf{Synthesis and verification.} The synthesizer receives the question, the selected draft or drafts and the numbered sources, each truncated to 1,400 characters. Its prompt treats the draft as an outline, requires concrete figures, programmes and targets to be taken from the sources, forbids content absent from them, and requires an {[}n{]} marker after each sentence or paragraph that uses a source. Because every candidate's sources are present, a fact retrieved only by a losing candidate can still reach the answer. The verifier then extracts the answer's claims and judges each against the original passages of all candidates, never against candidate drafts. Faithfulness is the share of supported claims: 0.8 or more is PASS, 0.5 to 0.8 PARTIAL, and below 0.5 FAIL. In C the verdict is only recorded. In G and I a FAIL replaces the answer with a fixed refusal, and a verification that cannot be parsed never triggers it.

\subsection{Worked example}

Figure 2 traces one multi-hop item (env\_hop\_05) through C in Qwen run 1. The question asks for trucks' share of Chungbuk's registered vehicles in 2020 and their share of on-road PM2.5 emissions, two facts reported in different documents. The rule for 대비 (contrast) marks it comparative, and an LLM call splits it into one sub-query per fact. Both sub-queries lost the year ``2020'' when list numbering was stripped from the model output, yet each local search still found its 2020 figure: 18.3\% in a 2022 transport carbon-neutrality roadmap and 77.0\% in a 2023 fine-dust strategy.

The selector scored the two drafts 8.5 and 9.5, because each answered only half of the question, and selected the second. The merged sources still carried the first draft's passages as {[}9{]}--{[}12{]}, so the synthesizer could join 18.3\% {[}9{]} with 77.0\% {[}1{]} and state the contrast that neither report makes. The verifier extracted three claims and found all of them supported. The judge scored this answer 2 of 2 for correctness, but the same item scored 0 in run 3, an instance of the run-to-run variation analysed in Section 6.1.

\begin{figure}
\centering
\includegraphics[width=0.82\linewidth,height=\textheight,keepaspectratio]{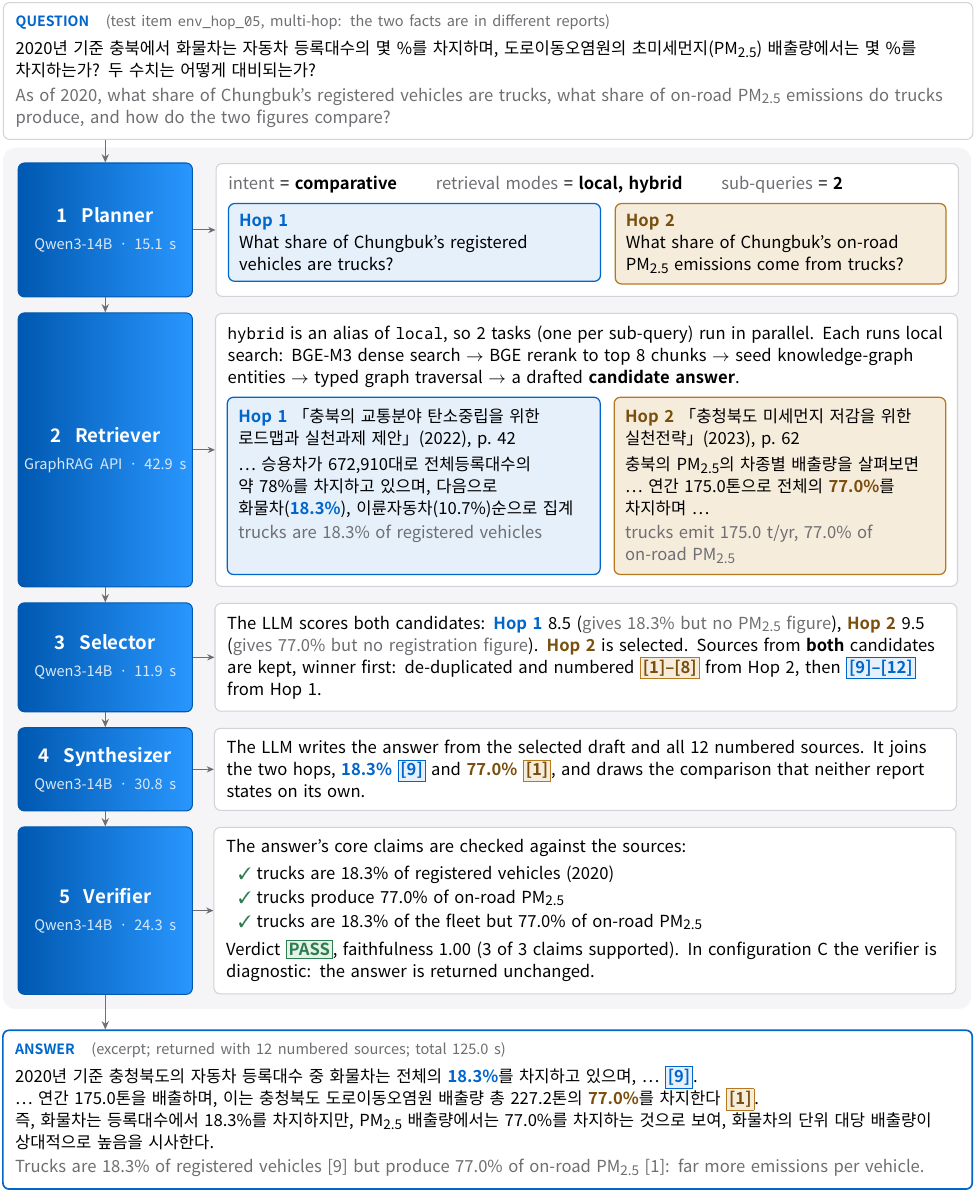}
\caption{One multi-hop question traced through MARO (item env\_hop\_05, configuration C, Qwen run 1). Hop 1 (blue) and hop 2 (amber) are retrieved by separate sub-queries from different reports; the selector keeps both candidates' sources, and the synthesizer cites one passage from each. All values, including stage latencies, are taken from the saved run record; Korean text is shown with English glosses.}
\end{figure}

\subsection{Model calls and settings}

Five conceptual stages do not mean five model calls per question. The harness recorded 345 calls per 100 questions for C in run 1. This count excludes the candidate-drafting call that the retrieval backend issues for each retrieval call (150 for C in run 1), so C's full total is about 495 (Section 6.7).

Original passages remain the primary evidence; graph context and candidate drafts are derived. Web search is disabled throughout.

Retrieval and generation use the following settings:

\begin{itemize}
\tightlist
\item
  \textbf{Retrieval (local mode):} dense search over-fetches 24 candidates, and bge-reranker-v2-m3 keeps the top 8.
\item
  \textbf{Context limits:} each passage is truncated to 1,400 characters, and the generation context is capped at 12,000 characters.
\item
  \textbf{Thinking mode:} disabled for all Qwen3-14B calls through the chat template.
\item
  \textbf{Temperature:} 0.1 for drafting, synthesis and planner decomposition; 0.0 for intent classification, candidate selection and verification.
\item
  \textbf{Top-p and precision:} top-p was not set by the client, so the serving default applied; serving precision was not recorded.
\end{itemize}

\section{Benchmark and compared systems}

The benchmark has 100 authored questions over 245 Korean regional policy-research documents, with gold evidence for every answerable item. All indexes and generated answers are frozen; later analyses use saved outputs only. Sections 4.5--4.7 define the configurations compared with MARO.

\subsection{Corpus and indexes}

\begin{longtable}[]{@{}ll@{}}
\caption{Corpus and index sizes.}\tabularnewline
\toprule\noalign{}
Component & Size \\
\midrule\noalign{}
\endfirsthead
\toprule\noalign{}
Component & Size \\
\midrule\noalign{}
\endhead
\bottomrule\noalign{}
\endlastfoot
Unique document texts & 245 \\
Dense index (BGE-M3, 1,024 dimensions) & 142,595 vectors \\
Knowledge graph (kg.graphml) & 114,304 nodes; 86,656 edges \\
LightRAG index & 14,102 internal chunks \\
\end{longtable}

Source PDFs were parsed with MinerU (hybrid backend, versions 3.1.11--3.4.0), with tables extracted as HTML. Each non-empty layout block (a paragraph, list or table, with running headers, footers and page numbers removed) is one retrieval unit tagged with its section path and page; no fixed-size windowing or overlap is applied. The dense index embeds each block's section path, content and caption with BGE-M3; indexed blocks have a median length of 71 characters (90th percentile 313).

\subsection{Questions}

The 100 questions comprise 30 factual, 30 comparison, 30 multi-hop and 10 unanswerable items. They are spread evenly over five source domains (20 each): culture and governance; environment, energy and agriculture; industry and technology; population, welfare and health; and regional development and transport. The questions were authored for this study, so the category balance defines a test instrument rather than an estimate of deployment query frequency.

\subsection{Gold evidence, unanswerable items and expert review}

The 90 answerable questions carry 214 evidence annotations referring to 204 distinct chunk IDs. Each record holds the document identifier, title, page and passage text. Questions, reference answers and evidence annotations were drafted with Claude Code (Anthropic) using claude-fable-5.1 under a written authoring brief. The model read each source document and ran exact substring searches over the parsed corpus to select evidence passages, which were copied verbatim by chunk ID; it did not query the BGE-M3 index, so the annotation does not favor dense retrieval. Reference answers were then expanded into full answers from the same cited passages.

Each unanswerable question has a reason and a verification note documenting the authors' corpus search. These notes support the unanswerable designation; they do not prove that no interpretation of the question is answerable.

An expert reviewer, a Korean professor of management information systems with 30 years of academic experience, then manually checked every question, reference answer and cited evidence passage for all 100 items and authorized their use before generation. The review record is retained with the study artifacts. This reference review is distinct from the generated-answer human study in Section 7.

\subsection{Isolation and provenance}

Reference answers, required points and gold evidence are available only to grading and retrieval measurement. The answering adapter receives question text alone. Dataset and run manifests record hashes, so later changes can be distinguished from the evaluated question set and corpus.

\subsection{Configurations}

We compare ten configurations: the full five-stage pipeline (C), six internal variants, a diagnostic-trace ablation (E), vanilla RAG (F) and LightRAG (J). The central comparison is C versus F. Because they differ in retrieval, prompting and orchestration, C\ensuremath{-}F measures two system bundles rather than the isolated effect of adding agents.

\begin{longtable}[]{@{}
  >{\raggedright\arraybackslash}p{(\linewidth - 4\tabcolsep) * \real{0.0556}}
  >{\raggedright\arraybackslash}p{(\linewidth - 4\tabcolsep) * \real{0.7222}}
  >{\raggedright\arraybackslash}p{(\linewidth - 4\tabcolsep) * \real{0.2222}}@{}}
\caption{Configurations compared. Generation runs are Qwen3-14B unless stated.}\tabularnewline
\toprule\noalign{}
\begin{minipage}[b]{\linewidth}\raggedright
ID
\end{minipage} & \begin{minipage}[b]{\linewidth}\raggedright
Configuration
\end{minipage} & \begin{minipage}[b]{\linewidth}\raggedright
Generation runs
\end{minipage} \\
\midrule\noalign{}
\endfirsthead
\toprule\noalign{}
\begin{minipage}[b]{\linewidth}\raggedright
ID
\end{minipage} & \begin{minipage}[b]{\linewidth}\raggedright
Configuration
\end{minipage} & \begin{minipage}[b]{\linewidth}\raggedright
Generation runs
\end{minipage} \\
\midrule\noalign{}
\endhead
\bottomrule\noalign{}
\endlastfoot
A & Vector retrieval with reranking; one retrieval call, drafted and synthesized (no planner, selector or verifier) & 3 \\
B & Fixed local graph-context retrieval; one retrieval call, drafted and synthesized (no planner, selector or verifier) & 3 \\
C & Full pipeline: planner, retriever, selector, synthesizer, diagnostic verifier & 3 Qwen + 3 EXAONE \\
D & C without planning & 3 \\
E & C with the diagnostic verifier removed from the saved record (identical answers) & Derived from C run 1 \\
F & Vanilla RAG: cosine top-k retrieval and one grounded generation call & 3 Qwen + 3 EXAONE \\
G & C with a verification-triggered refusal guard & 3 \\
H & C with graph-linked chunks added to the candidate pool before joint reranking & 3 \\
I & C with both the guard and graph-linked chunk expansion & 3 \\
J & LightRAG, hybrid mode (external baseline) & 3 \\
\end{longtable}

F retrieves the top 8 blocks by raw cosine similarity, without reranking, graph expansion or community context, and answers in one generation call (temperature 0.1, at most 1,024 tokens) under the same per-passage and total context limits. Its prompt (vanilla-v1, Supplement S1) instructs the model to answer only from the retrieved content, to cite report title and page, and to reply ``검색된 자료에서 확인할 수 없습니다'' (``this cannot be confirmed in the retrieved material'') when the content does not contain the answer.

E is not an independently generated baseline. Its answer strings are identical to run-1 C for all 100 questions; it differs only in the trace the judge sees. This makes E a natural control for judge-packet sensitivity (Section 6.3).

H and I add graph-linked chunks to the retrieval pool before joint reranking. This differs from the graph context in the local path, where graph information enriches results without competing for rank.

\subsection{Generators}

All Qwen experiments use qwen3-14b (Yang et al., 2025) as the generator for every stage and configuration. To test generator dependence, C and F were also generated three times with EXAONE 4.0 32B (Bae et al., 2025), a Korean-native model from a different family than both the Qwen generator and the Gemma judge. It ran as an 8-bit MLX quantization (lmstudio-community/EXAONE-4.0-32B-MLX-8bit) with thinking disabled, on the same frozen dataset, indexes and judge. A fourth EXAONE generation, served in FP8 on a different server, is reported as a supplementary single run. F's retrieval is deterministic, so its source lists contain the same chunks across generators and runs; C's retrieval depends on generated plans.

\subsection{LightRAG baseline}

J uses its own index and hybrid retrieval. The frozen build manifest records a 1,200-token chunk size, query top\_k = 40 and chunk\_top\_k = 20, consistent with the defaults of the installed LightRAG 1.5.7. The index and these settings were frozen before J's answers were generated. Index construction and all saved generation calls used the same qwen3-14b model identifier. Each LightRAG chunk maps to 10.74 original blocks on average, so recall@k for J counts different units than for the other systems (Section 5.5).

\section{Evaluation protocol}

We vary six evaluation choices on the same saved outputs and report how each changes the C-versus-F verdict. No answer, index or original grade was regenerated or overwritten.

Formally, a protocol \ensuremath{\pi} = (g, R, j, P, k) fixes the generator g, the set of generation runs R, the judge packet j, the scoring population P and the criterion k; together these determine every score from which a contrast is computed. Four of the six choices below vary \ensuremath{\pi} directly: repetition varies R, generator family g, judge input j and scoring population P. The other two, abstention coding and retrieval normalization, change the statistic that is computed rather than the scores, and are analysed separately (Sections 6.5 and 6.6). Section 5.6 defines how a contrast is summarized over a family of protocols. Figure 3 shows the framework, from the saved outputs to the profile of each contrast.

\begin{longtable}[]{@{}
  >{\raggedright\arraybackslash}p{(\linewidth - 6\tabcolsep) * \real{0.2500}}
  >{\raggedright\arraybackslash}p{(\linewidth - 6\tabcolsep) * \real{0.2500}}
  >{\raggedright\arraybackslash}p{(\linewidth - 6\tabcolsep) * \real{0.2500}}
  >{\raggedright\arraybackslash}p{(\linewidth - 6\tabcolsep) * \real{0.2500}}@{}}
\caption{The six evaluation choices varied on the same saved outputs, and the results subsection reporting each.}\tabularnewline
\toprule\noalign{}
\begin{minipage}[b]{\linewidth}\raggedright
Evaluation choice
\end{minipage} & \begin{minipage}[b]{\linewidth}\raggedright
Setting A
\end{minipage} & \begin{minipage}[b]{\linewidth}\raggedright
Setting B
\end{minipage} & \begin{minipage}[b]{\linewidth}\raggedright
Results
\end{minipage} \\
\midrule\noalign{}
\endfirsthead
\toprule\noalign{}
\begin{minipage}[b]{\linewidth}\raggedright
Evaluation choice
\end{minipage} & \begin{minipage}[b]{\linewidth}\raggedright
Setting A
\end{minipage} & \begin{minipage}[b]{\linewidth}\raggedright
Setting B
\end{minipage} & \begin{minipage}[b]{\linewidth}\raggedright
Results
\end{minipage} \\
\midrule\noalign{}
\endhead
\bottomrule\noalign{}
\endlastfoot
Generation repetition & Run 1 only & Three runs, averaged within item & 6.1 \\
Generator family & Qwen3-14B & EXAONE 4.0 32B & 6.2 \\
Judge input & Trace-aware packet & Answer-only common packet & 6.3 \\
Scoring population & 90 answerable questions & All 100 questions & 6.4 \\
Abstention coding & Binary (judge field) & Refusal / hedge / answer & 6.5 \\
Retrieval normalization & Source-count recall@k & Gold-evidence recall at matched character budgets & 6.6 \\
\end{longtable}

\begin{figure}
\centering
\includegraphics[width=1\linewidth,height=\textheight,keepaspectratio]{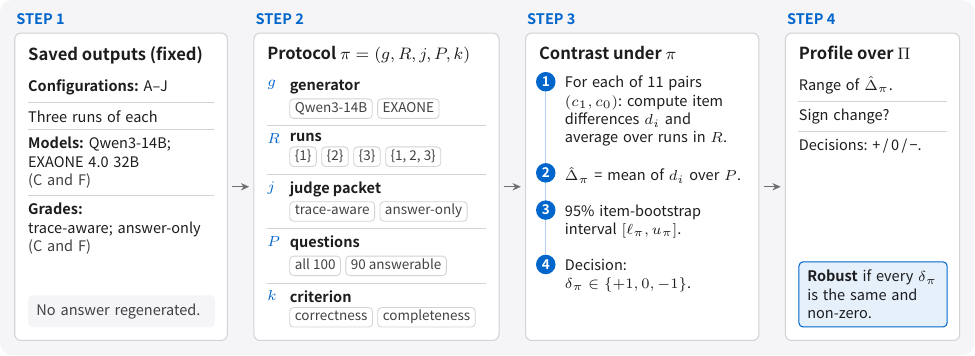}
\caption{Protocol-sensitivity framework. A protocol \ensuremath{\pi} fixes the generator g, the run set R, the judge packet j, the scoring population P and the criterion k. Each of the eleven paired contrasts is recomputed under every protocol the saved outputs support, and its profile over the protocol family \ensuremath{\Pi} reports the range of estimates, sign changes and decisions (Section 5.6). No answer is regenerated.}
\end{figure}

\subsection{Generation runs}

Run 1 contains all ten configurations. Runs 2 and 3 contain the nine independently generated ones (A, B, C, D, F, G, H, I, J); E exists only for run 1. This gives 2,800 saved Qwen records, of which 2,700 are independent generations, all with successful status. The EXAONE study adds 600 records (C and F, three runs) and the supplementary FP8 run adds 200.

\subsection{Judge protocols}

\textbf{Answer-quality criteria.} The judge, in both protocols below, and the human reviewers (Section 7) score each answer on the same criteria, each on a 0--2 scale: 0 = main failure, 1 = partial success or material omission, 2 = success (Table 5). Every criterion is judged against the question's reference answer, required points and gold passages, never against outside knowledge, and neither length nor fluency earns credit. A criterion that cannot be judged from the supplied material is marked cannot\_assess, and one that does not apply is marked not\_applicable; both are excluded from that criterion's mean. The judge rubric spells out temporal accuracy and readability in full; for correctness, completeness and evidence support it gives the shared anchors only, and the descriptions in Table 5 follow the reviewer guide, which uses the same anchors.

\begin{longtable}[]{@{}
  >{\raggedright\arraybackslash}p{(\linewidth - 4\tabcolsep) * \real{0.1818}}
  >{\raggedright\arraybackslash}p{(\linewidth - 4\tabcolsep) * \real{0.5455}}
  >{\raggedright\arraybackslash}p{(\linewidth - 4\tabcolsep) * \real{0.2727}}@{}}
\caption{Answer-quality criteria, scored 0--2 by the judge and by human reviewers.}\tabularnewline
\toprule\noalign{}
\begin{minipage}[b]{\linewidth}\raggedright
Criterion
\end{minipage} & \begin{minipage}[b]{\linewidth}\raggedright
What it measures
\end{minipage} & \begin{minipage}[b]{\linewidth}\raggedright
Scored on
\end{minipage} \\
\midrule\noalign{}
\endfirsthead
\toprule\noalign{}
\begin{minipage}[b]{\linewidth}\raggedright
Criterion
\end{minipage} & \begin{minipage}[b]{\linewidth}\raggedright
What it measures
\end{minipage} & \begin{minipage}[b]{\linewidth}\raggedright
Scored on
\end{minipage} \\
\midrule\noalign{}
\endhead
\bottomrule\noalign{}
\endlastfoot
Correctness & Whether what the answer states is right, checked against the reference answer and gold passages. A fluent but wrong answer scores 0. & All questions; on unanswerable ones a clean refusal can earn full credit \\
Completeness & How many of the question's required points the answer covers. & Answerable questions \\
Evidence support & Whether the answer's claims are backed by the evidence passages; fluent unsupported claims earn no credit. & Answerable questions \\
Temporal accuracy & Whether each figure, policy or event is tied to the right year or period, and whether the answer covers the period asked about. A value from the wrong year scores 0 even if the number appears in a source; a missing, vague or partly wrong period scores 1. & Answerable questions with a time element \\
Readability (optional) & Structure, clarity and absence of redundancy for a policy researcher. Never substitutes for correctness. & Substantive answers only \\
Refusal verdict (yes/no) & Whether the answer declined, rather than supplying figures for an adjacent year or region. & Unanswerable questions \\
\end{longtable}

In trace-aware grading the judge could also check claims against the system's returned sources; in answer-only grading and for human reviewers, evidence support is judged against the gold passages only. The judge's refusal verdict (abstention\_correct) proved unreliable and is not used; refusal statistics come from the coding in Section 5.4.

\textbf{Trace-aware grading (original protocol).} The judge of record is google/gemma-4-31B-it with frozen judge-v5 instructions. It scores correctness, completeness, evidence support and temporal accuracy on a 0--2 scale; readability is optional. Non-numeric cannot\_assess and not\_applicable values are excluded criterion by criterion. The packet contains the question, reference answer, required points, gold passages, final answer, returned sources, and stage inputs and outputs. It omits the configuration identifier, but stage roles and traces reveal the architecture.

The default temperature is 0. One run-1 J packet (env\_unans\_01\_\_J) needed a recorded temperature-0.2 deviation after reproducible malformed output; the corresponding run-2 J item remains ungraded. In total, 2,799 of 2,800 Qwen grades succeeded.

\textbf{Answer-only grading (sensitivity audit).} After observing different grades for identical C/E answer strings, we re-graded the 600 saved Qwen C and F answers. The common packet contains the question and its category, reference answer, required points, acceptable variants, numerical tolerance, gold passage identifiers and text, and the unchanged final answer. It excludes configuration identifiers, system-specific source lists, intermediate traces, call counts and latency, so identical answers to the same question receive identical packets.

The frozen judge-v5-answer-only-v1 rubric keeps the original answer-score definitions and removes agent-role scoring. Temperature is 0, maximum output 2,048 tokens, with structured JSON enforced. The endpoint advertised the same model identifier as the original judge, but its weights, quantization and serving equivalence were not verified. The audit produced 599 valid grades; run 2's C answer to cul\_unans\_01 failed twice with malformed JSON and is left ungraded.

The audit changes three things at once: trace visibility, source-list visibility and the serving endpoint. It is therefore a grading-protocol sensitivity check, not a causal estimate of trace influence. The C/E comparison (Section 6.3) is the cleaner test of trace influence, because only the packet differs.

\textbf{EXAONE grading.} The main EXAONE study has 595 of 600 successful grades and the FP8 run 198 of 200. All seven failures are C answers to two unanswerable questions (cul\_unans\_01, reg\_unans\_01) with the same malformed output: in the main study, reg\_unans\_01 in all three runs and cul\_unans\_01 in runs 1 and 2; in the FP8 run, both. F scores 2 on both questions in every run, so each missing C\ensuremath{-}F difference lies between \ensuremath{-}2 and 0. Because reg\_unans\_01 has no C grade in any main run, the item-averaged estimate in Table 9 covers N = 99 questions. Setting the missing C scores to 0 or to 2 gives C\ensuremath{-}F = \ensuremath{-}0.137 {[}\ensuremath{-}0.303, +0.030{]} or \ensuremath{-}0.103 {[}\ensuremath{-}0.260, +0.053{]} (N = 100); every admissible completion leaves the estimate negative with an interval that includes zero. For the FP8 run the corresponding bounds are \ensuremath{-}0.050 {[}\ensuremath{-}0.220, +0.120{]} and \ensuremath{-}0.010 {[}\ensuremath{-}0.170, +0.150{]}.

\subsection{Scoring populations}

We report two machine-correctness summaries. Answerable-only means use the 90 answerable questions. Paired contrasts use all 100 questions, including the judge's correctness score on unanswerable items, where a clean refusal can earn full credit. The two summaries answer different questions and their numbers are not interchangeable.

\subsection{Abstention coding}

Each answer text is coded into three categories by a fixed keyword rule, applied by script:

\begin{itemize}
\tightlist
\item
  \textbf{Clean refusal:} declines the requested answer without supplying a substitute. Operationally, the answer contains the refusal phrase 확인할 수 없습니다 (``cannot be confirmed'') and is shorter than 400 characters.
\item
  \textbf{Hedge:} acknowledges missing evidence but proceeds with adjacent or insufficient information. Operationally, the answer contains the refusal phrase but is 400 characters or longer, or contains one of seven gap phrases (for example 명시되어 있지 않, 찾을 수 없, 알 수 없).
\item
  \textbf{Answer:} attempts a substantive answer without a qualifying refusal.
\end{itemize}

Coding is reported separately for answerable and unanswerable items; a clean refusal on an answerable item is a false refusal. The coding is separate from correctness and from the judge's abstention\_correct field, which agreed with human refusal verdicts on only 45 of 80 cases (Section 7.3). Because the rule is deterministic and sees only the answer text, the coding is blind to configuration by construction and exactly reproducible. As validation, we compared the rule's clean-refusal decision with the human reviewers' refusal verdicts on the 80 run-1 unanswerable ratings: they agree on 67 of 80 (Cohen's \ensuremath{\kappa} = 0.675, 95\% bootstrap interval 0.50--0.83 over ratings). This checks the refusal boundary only; no human coder checked the hedge/answer boundary. A, B, D, J and the EXAONE runs are not coded.

\subsection{Retrieval measurement}

The main retrieval measure asks a simple question: how much of the annotated gold evidence appears in the first 3,000 characters of the sources a system returns? We call this gold-evidence recall at a 3,000-character budget. We also report it at 2,000, 5,000 and 10,000 characters, and with no limit at all (the all-source ceiling). Formally, for question i with gold IDs G\_i, source texts t\_ij and source IDs c\_ij:

\[R_i(B) = \frac{\left| G_i \cap \left\{ c_{ij} : \sum_{k \le j} |t_{ik}| \le B,\ |t_{ij}| > 0 \right\} \right|}{|G_i|}\]

Characters are counted over source content only (Python Unicode counts, including whitespace and HTML), in the order the system returned the sources; headers, titles, citation markers and prompt instructions are not counted. A gold passage is credited only if its complete, non-empty text fits within the budget. A passage cut off by the limit gets no credit, so the measure is conservative. The budgets were chosen after inspecting the data. The 3,000-character point is close to the amount of text the internal systems return: in run 1, C's first eight sources have a median of 2,167 characters on answerable questions.

Only answerable questions with gold evidence and a successful generation are scored. Outputs whose judge grade failed are still included, because this measure does not use the judge. Recall is averaged over questions within each run, then over the three runs. We also report recall@5, recall@8 and MRR, but only as secondary measures: they count sources, and one source is a different amount of text in each system. Recall@k is the share of a question's gold passages found among the first k returned sources, and MRR is the reciprocal rank of the first gold passage returned (0 if none is).

\subsection{Statistical analysis}

For configurations c1 and c0, let s\_irc be the correctness score for question i in run r. The paired item difference averages over runs R\_i with valid grades on both sides, and the effect averages over questions:

\[d_i = \frac{1}{|R_i|} \sum_{r \in R_i} \left( s_{irc_1} - s_{irc_0} \right), \qquad \hat{\Delta} = \frac{1}{N} \sum_{i=1}^{N} d_i\]

Thus each question contributes once, and repeated runs reduce noise within question. Percentile intervals resample questions 10,000 times (seed 20261020). These intervals condition on the saved generations; with three runs they do not fully characterize future run-to-run variation.

Human-rating intervals use the same item bootstrap (10,000 question resamples, seed 20261020). All intervals are nominal 95\% intervals, not adjusted for multiple contrasts or criteria. Correctness is the focal endpoint but was not preregistered.

An interval that includes zero does not establish equivalence. We therefore report which effect sizes each interval rules out and, against a smallest effect of interest of \ensuremath{\pm}0.15 on the 0--2 scale, two one-sided tests based on bootstrap 90\% intervals. These establish equivalence for G\ensuremath{-}C, H\ensuremath{-}C and I\ensuremath{-}C (p \ensuremath{\leq} 0.003) and, marginally, for C\ensuremath{-}B and B\ensuremath{-}F (p \ensuremath{\approx} 0.05); C\ensuremath{-}D lies on the margin. The principal C\ensuremath{-}F contrast is not equivalent under trace-aware grading (90\% interval {[}\ensuremath{-}0.057, +0.177{]}) but is under answer-only grading ({[}\ensuremath{-}0.133, +0.110{]}, p = 0.034).

\textbf{Protocol-sensitivity profile.} Let \ensuremath{\hat{\Delta}}\_\ensuremath{\pi}(c1, c0) be the contrast above computed under protocol \ensuremath{\pi}, with the runs R\_i drawn from \ensuremath{\pi}'s run set and the mean taken over its population P, and let {[}\ensuremath{\ell}\_\ensuremath{\pi}, u\_\ensuremath{\pi}{]} be its 95\% interval. Its decision \ensuremath{\delta}\_\ensuremath{\pi} is +1 if the interval lies above zero, \ensuremath{-}1 if it lies below zero and 0 otherwise. For a family \ensuremath{\Pi} of protocols that the saved data support, the profile of a contrast reports the range of \ensuremath{\hat{\Delta}}\_\ensuremath{\pi} over \ensuremath{\Pi}, whether the point estimates take both signs, and how many protocols give each decision. A verdict is protocol-robust over \ensuremath{\Pi} when every protocol gives the same non-zero decision:

\[\delta_\pi = \operatorname{sign}(\ell_\pi)\,\mathbb{1}\left[\ell_\pi u_\pi > 0\right], \qquad \text{robust}(\Pi) \iff \exists\, \delta \neq 0 : \delta_\pi = \delta \;\; \forall \pi \in \Pi\]

For correctness, \ensuremath{\Pi} crosses the run set (each single run and the three-run pool) with the scoring population (all 100 or the 90 answerable questions). Completeness is scored on the 90 answerable questions only. C\ensuremath{-}F is the only contrast generated with both generators and graded with both judge packets, so its family has 24 correctness and 12 completeness protocols; every other contrast has 8 and 4. The profile describes how far a conclusion depends on the protocol. It is not a hypothesis test: protocols share questions and, within a generator, answers, so their decisions are not independent.

Resource measures are model-call counts and recorded prompt and completion tokens. Wall-clock latency was measured on a shared server under uncontrolled load and is not used for comparison. Offline index construction is excluded.

\section{Results: how each evaluation choice changes the verdict}

MARO's answers are more complete than vanilla RAG's under every protocol, but its correctness advantage depends on the protocol. Four of the six evaluation choices shrink or reverse at least one architecture contrast, and two change what the comparison can see. Only completeness contrasts hold under every protocol: C, A and B over vanilla RAG, and C over LightRAG, so the simplified paths A and B share MARO's completeness gain over vanilla RAG. Table 6 summarizes the effect of each choice; Sections 6.1--6.8 give the details, and Section 6.9 profiles all eleven contrasts. Figure 4 shows the C\ensuremath{-}F correctness and completeness contrasts under the main protocols.

\begin{longtable}[]{@{}
  >{\raggedright\arraybackslash}p{(\linewidth - 6\tabcolsep) * \real{0.2500}}
  >{\raggedright\arraybackslash}p{(\linewidth - 6\tabcolsep) * \real{0.2500}}
  >{\raggedright\arraybackslash}p{(\linewidth - 6\tabcolsep) * \real{0.2500}}
  >{\raggedright\arraybackslash}p{(\linewidth - 6\tabcolsep) * \real{0.2500}}@{}}
\caption{How each evaluation choice changes the conclusion.}\tabularnewline
\toprule\noalign{}
\begin{minipage}[b]{\linewidth}\raggedright
Evaluation choice
\end{minipage} & \begin{minipage}[b]{\linewidth}\raggedright
Narrow reading
\end{minipage} & \begin{minipage}[b]{\linewidth}\raggedright
Fuller reading
\end{minipage} & \begin{minipage}[b]{\linewidth}\raggedright
Effect on verdict
\end{minipage} \\
\midrule\noalign{}
\endfirsthead
\toprule\noalign{}
\begin{minipage}[b]{\linewidth}\raggedright
Evaluation choice
\end{minipage} & \begin{minipage}[b]{\linewidth}\raggedright
Narrow reading
\end{minipage} & \begin{minipage}[b]{\linewidth}\raggedright
Fuller reading
\end{minipage} & \begin{minipage}[b]{\linewidth}\raggedright
Effect on verdict
\end{minipage} \\
\midrule\noalign{}
\endhead
\bottomrule\noalign{}
\endlastfoot
Generation repetition & Run 1: C leads A, B and D by 0.14--0.19 and F by 0.150; G, H and I trail C & Runs 2--3: C's leads fall to 0.00--0.11; G, H and I change sign in both later runs & Shrinks or reverses \\
Generator family & Qwen: C leads F by about 0.14 on answerable questions & EXAONE: C trails F by 0.026 on answerable questions and in all three runs overall & Reverses the point estimate \\
Judge input & Trace-aware: C\ensuremath{-}F = +0.060; identical C and E answers score 1.544 and 1.500 & Answer-only: C\ensuremath{-}F = \ensuremath{-}0.010 & Shrinks to near zero \\
Scoring population & Answerable-only: A and B score about 0.10 above F & All questions: A\ensuremath{-}F = \ensuremath{-}0.037, B\ensuremath{-}F = \ensuremath{-}0.010 & Reverses \\
Abstention coding & Binary judge field agrees with humans on 45 of 80 refusal verdicts & C hedges and never cleanly refuses; F refuses 18--19 of 90 answerable questions per run & Reveals hidden behavior \\
Retrieval normalization & Recall@5: J = 0.147 versus C = 0.481 & All-source ceiling: J = 0.594 exceeds C = 0.560; J's deficit is evidence density & Changes the explanation \\
Criterion & Correctness: no all-question interval excludes zero; answerable-only C\ensuremath{-}F does (+0.141) & Completeness: +0.226 (answer-only) and +0.367 (human) favor C & Survives every protocol \\
\end{longtable}

\begin{figure}
\centering
\includegraphics[width=1\linewidth,height=\textheight,keepaspectratio]{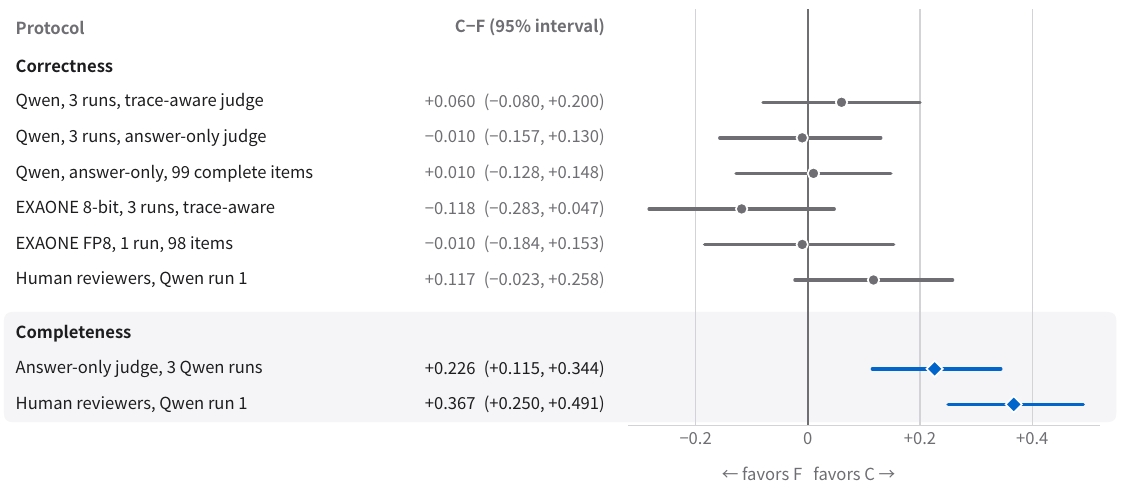}
\caption{Six correctness estimates under different runs, generators and judges all include zero, while both completeness estimates exclude it. Correctness covers all 100 questions; restricted to the 90 answerable questions, the Qwen C\ensuremath{-}F interval excludes zero (+0.141 {[}+0.015, +0.274{]}, Section 6.4).}
\end{figure}

\subsection{Generation repetition}

Repeating generation shrank every run-1 contrast involving C. C's answerable-question mean dropped from 1.544 in run 1 to 1.400 and 1.433 in runs 2 and 3, the largest spread of any configuration (Table 7).

\begin{longtable}[]{@{}llllll@{}}
\caption{Machine correctness on answerable questions (trace-aware grading, 0--2 scale, Qwen).}\tabularnewline
\toprule\noalign{}
Config & Run 1 & Run 2 & Run 3 & Mean & Spread \\
\midrule\noalign{}
\endfirsthead
\toprule\noalign{}
Config & Run 1 & Run 2 & Run 3 & Mean & Spread \\
\midrule\noalign{}
\endhead
\bottomrule\noalign{}
\endlastfoot
A & 1.433 & 1.422 & 1.400 & 1.419 & 0.033 \\
B & 1.400 & 1.400 & 1.433 & 1.411 & 0.033 \\
C & 1.544 & 1.400 & 1.433 & 1.459 & 0.144 \\
D & 1.411 & 1.400 & 1.411 & 1.407 & 0.011 \\
F & 1.311 & 1.311 & 1.333 & 1.319 & 0.022 \\
G & 1.356 & 1.344 & 1.422 & 1.374 & 0.078 \\
H & 1.400 & 1.456 & 1.444 & 1.433 & 0.056 \\
I & 1.367 & 1.400 & 1.478 & 1.415 & 0.111 \\
J & 1.211 & 1.211 & 1.244 & 1.222 & 0.033 \\
\end{longtable}

Three runs are too few for a stable variance estimate, and the larger spread does not show that added stages cause variance: G, H and I add stages to C and vary less. The simplified paths A and B score within C's run-to-run range; C clearly exceeds both only in run 1 (in run 3 it leads A by 0.033 on answerable questions and ties B).

\begin{longtable}[]{@{}
  >{\raggedright\arraybackslash}p{(\linewidth - 6\tabcolsep) * \real{0.2500}}
  >{\raggedright\arraybackslash}p{(\linewidth - 6\tabcolsep) * \real{0.2500}}
  >{\raggedright\arraybackslash}p{(\linewidth - 6\tabcolsep) * \real{0.2500}}
  >{\raggedright\arraybackslash}p{(\linewidth - 6\tabcolsep) * \real{0.2500}}@{}}
\caption{Paired machine correctness over all 100 questions (trace-aware grading, Qwen).}\tabularnewline
\toprule\noalign{}
\begin{minipage}[b]{\linewidth}\raggedright
Contrast
\end{minipage} & \begin{minipage}[b]{\linewidth}\raggedright
Run 1 / 2 / 3
\end{minipage} & \begin{minipage}[b]{\linewidth}\raggedright
Item-averaged difference
\end{minipage} & \begin{minipage}[b]{\linewidth}\raggedright
Nominal 95\% interval
\end{minipage} \\
\midrule\noalign{}
\endfirsthead
\toprule\noalign{}
\begin{minipage}[b]{\linewidth}\raggedright
Contrast
\end{minipage} & \begin{minipage}[b]{\linewidth}\raggedright
Run 1 / 2 / 3
\end{minipage} & \begin{minipage}[b]{\linewidth}\raggedright
Item-averaged difference
\end{minipage} & \begin{minipage}[b]{\linewidth}\raggedright
Nominal 95\% interval
\end{minipage} \\
\midrule\noalign{}
\endhead
\bottomrule\noalign{}
\endlastfoot
C\ensuremath{-}F & +0.150 / 0.000 / +0.030 & +0.060 & {[}\ensuremath{-}0.080, +0.200{]} \\
C\ensuremath{-}A & +0.140 / +0.040 / +0.110 & +0.097 & {[}\ensuremath{-}0.017, +0.213{]} \\
C\ensuremath{-}B & +0.190 / 0.000 / +0.020 & +0.070 & {[}\ensuremath{-}0.023, +0.163{]} \\
C\ensuremath{-}D & +0.140 / +0.020 / +0.040 & +0.067 & {[}\ensuremath{-}0.030, +0.167{]} \\
A\ensuremath{-}F & +0.010 / \ensuremath{-}0.040 / \ensuremath{-}0.080 & \ensuremath{-}0.037 & {[}\ensuremath{-}0.213, +0.140{]} \\
B\ensuremath{-}F & \ensuremath{-}0.040 / 0.000 / +0.010 & \ensuremath{-}0.010 & {[}\ensuremath{-}0.170, +0.157{]} \\
G\ensuremath{-}C & \ensuremath{-}0.120 / +0.010 / +0.030 & \ensuremath{-}0.027 & {[}\ensuremath{-}0.107, +0.060{]} \\
H\ensuremath{-}C & \ensuremath{-}0.150 / +0.030 / +0.010 & \ensuremath{-}0.037 & {[}\ensuremath{-}0.117, +0.040{]} \\
I\ensuremath{-}C & \ensuremath{-}0.100 / +0.060 / +0.080 & +0.013 & {[}\ensuremath{-}0.073, +0.103{]} \\
I\ensuremath{-}F & +0.050 / +0.060 / +0.110 & +0.073 & {[}\ensuremath{-}0.060, +0.210{]} \\
J\ensuremath{-}C & \ensuremath{-}0.340 / \ensuremath{-}0.172 / \ensuremath{-}0.210 & \ensuremath{-}0.247 & {[}\ensuremath{-}0.417, \ensuremath{-}0.077{]} \\
J\ensuremath{-}F & \ensuremath{-}0.190 / \ensuremath{-}0.172 / \ensuremath{-}0.180 & \ensuremath{-}0.187 & {[}\ensuremath{-}0.373, \ensuremath{-}0.003{]} \\
\end{longtable}

In run 1, C led A, B and D by 0.14--0.19, enough to read as a clear benefit of orchestration and planning. In runs 2 and 3 those leads fell to 0.00--0.11. G, H and I trailed C in run 1 and led it in both later runs. For every contrast involving C, run 1 shows the largest magnitude of the three runs. None of the six internal contrasts with C excludes zero.

The C\ensuremath{-}F interval excludes advantages for C larger than +0.200 and disadvantages larger than \ensuremath{-}0.080; it establishes neither superiority nor equivalence. Even in run 1 alone, the C\ensuremath{-}F interval included zero. J scores below C in all three runs, with an interval clearly below zero. J\ensuremath{-}F is negative in every run, but its interval ends just below zero (upper bound \ensuremath{-}0.003); given the number of contrasts and one ungraded item, we report the consistent direction without claiming a difference.

MARO's prompts, routing rules and thresholds were last changed on 14 September 2026, before the questions existed. The 100 questions were drafted on 28--29 September and frozen on 30 September at 07:25 UTC. Run 1 was generated from the unmodified committed code between 08:10 and 10:30 UTC the same day, and no C answer required a retry. MARO was therefore not tuned on these questions through committed changes, and the authors confirm that no informal testing against the chat interface took place before run 1.

\subsection{Generator family}

With EXAONE 4.0 32B as the generator, C scores below F in all three runs, on both summaries (Table 9). The Qwen answerable-only advantage of about 0.14 for C does not reappear.

\begin{longtable}[]{@{}
  >{\raggedright\arraybackslash}p{(\linewidth - 10\tabcolsep) * \real{0.1667}}
  >{\raggedright\arraybackslash}p{(\linewidth - 10\tabcolsep) * \real{0.1667}}
  >{\raggedright\arraybackslash}p{(\linewidth - 10\tabcolsep) * \real{0.1667}}
  >{\raggedright\arraybackslash}p{(\linewidth - 10\tabcolsep) * \real{0.1667}}
  >{\raggedright\arraybackslash}p{(\linewidth - 10\tabcolsep) * \real{0.1667}}
  >{\raggedright\arraybackslash}p{(\linewidth - 10\tabcolsep) * \real{0.1667}}@{}}
\caption{EXAONE 4.0 32B correctness, C versus F (trace-aware grading).}\tabularnewline
\toprule\noalign{}
\begin{minipage}[b]{\linewidth}\raggedright
Measure
\end{minipage} & \begin{minipage}[b]{\linewidth}\raggedright
Run 1
\end{minipage} & \begin{minipage}[b]{\linewidth}\raggedright
Run 2
\end{minipage} & \begin{minipage}[b]{\linewidth}\raggedright
Run 3
\end{minipage} & \begin{minipage}[b]{\linewidth}\raggedright
Item-averaged
\end{minipage} & \begin{minipage}[b]{\linewidth}\raggedright
Nominal 95\% interval
\end{minipage} \\
\midrule\noalign{}
\endfirsthead
\toprule\noalign{}
\begin{minipage}[b]{\linewidth}\raggedright
Measure
\end{minipage} & \begin{minipage}[b]{\linewidth}\raggedright
Run 1
\end{minipage} & \begin{minipage}[b]{\linewidth}\raggedright
Run 2
\end{minipage} & \begin{minipage}[b]{\linewidth}\raggedright
Run 3
\end{minipage} & \begin{minipage}[b]{\linewidth}\raggedright
Item-averaged
\end{minipage} & \begin{minipage}[b]{\linewidth}\raggedright
Nominal 95\% interval
\end{minipage} \\
\midrule\noalign{}
\endhead
\bottomrule\noalign{}
\endlastfoot
C, answerable mean & 1.300 & 1.322 & 1.333 & 1.319 & --- \\
F, answerable mean & 1.322 & 1.356 & 1.356 & 1.344 & --- \\
C\ensuremath{-}F, all questions & \ensuremath{-}0.082 & \ensuremath{-}0.112 & \ensuremath{-}0.121 & \ensuremath{-}0.118 & {[}\ensuremath{-}0.283, +0.047{]} \\
\end{longtable}

The item-averaged C\ensuremath{-}F (\ensuremath{-}0.118, N = 99) is more negative than runs 1 and 2 and close to run 3 (\ensuremath{-}0.121). Item averaging keeps cul\_unans\_01, whose only graded run (run 3) gives C\ensuremath{-}F = \ensuremath{-}2, while the run-1 and run-2 summaries drop it; reg\_unans\_01 has no C grade in any run and is excluded throughout. Bounding the missing scores between 0 and 2 places C\ensuremath{-}F between \ensuremath{-}0.137 and \ensuremath{-}0.103 (Section 5.2). The supplementary FP8 run gives C\ensuremath{-}F = \ensuremath{-}0.010 {[}\ensuremath{-}0.184, +0.153{]} on 98 paired questions, although there C leads F on answerable questions (1.378 against 1.333).

C's EXAONE run-to-run spread on answerable questions (0.033) is much smaller than its Qwen spread (0.144), so the large Qwen variability is not a fixed property of the pipeline. The cost overhead is similar across generators: 339--340 model calls per 100 questions for C against F's 100, and about 706,000 against 224,000 prompt tokens. C's 3,000-character evidence recall with EXAONE is 0.496, against F's 0.565 from the same deterministic source sets. Refusal behavior under EXAONE has not been coded.

Ferrazzi et al.~(2026) found the gap between agentic and enhanced RAG largely stable across model sizes. Here the direction of the C\ensuremath{-}F point estimate changed with the generator family, so a single-generator result would have supported a different conclusion.

\subsection{Judge input}

What the judge saw changed scores for identical answers. In run 1, C and E return identical answer strings for all 100 questions; only the trace in the judge packet differs. Their answerable-question means are 1.544 (C) and 1.500 (E). Because the text is the same, this gap measures grading sensitivity to packet content, not any effect of verification on the answer. Of the 100 pairs, 95 received the same correctness score; the other five all favored C (four answerable items by one point and one unanswerable item by two), which accounts for the entire answerable-question gap. The answer-only audit did not re-grade E.

The answer-only audit tests the principal comparison under a common packet. C\ensuremath{-}F correctness becomes \ensuremath{-}0.010 {[}\ensuremath{-}0.157, +0.130{]}, against +0.060 under trace-aware grading. Per-run contrasts are 0.000, +0.051 and \ensuremath{-}0.060 (run 2 uses 99 matched questions). Across item-averaged scores, C wins 23 questions, ties 56 and loses 21. Restricting to the 99 questions with all three matched runs gives +0.010 {[}\ensuremath{-}0.128, +0.148{]}.

Run 1 is the most sensitive to the protocol. C's all-question correctness mean falls from 1.530 to 1.370, while F's falls only from 1.380 to 1.370. Because traces, source lists and the endpoint changed together, this cannot be attributed to traces alone or read as systematic favoritism toward multi-stage systems.

\begin{longtable}[]{@{}
  >{\raggedright\arraybackslash}p{(\linewidth - 10\tabcolsep) * \real{0.1667}}
  >{\raggedright\arraybackslash}p{(\linewidth - 10\tabcolsep) * \real{0.1667}}
  >{\raggedright\arraybackslash}p{(\linewidth - 10\tabcolsep) * \real{0.1667}}
  >{\raggedright\arraybackslash}p{(\linewidth - 10\tabcolsep) * \real{0.1667}}
  >{\raggedright\arraybackslash}p{(\linewidth - 10\tabcolsep) * \real{0.1667}}
  >{\raggedright\arraybackslash}p{(\linewidth - 10\tabcolsep) * \real{0.1667}}@{}}
\caption{Answer-only grading of saved C and F answers (three Qwen runs, 0--2 scale).}\tabularnewline
\toprule\noalign{}
\begin{minipage}[b]{\linewidth}\raggedright
Criterion
\end{minipage} & \begin{minipage}[b]{\linewidth}\raggedright
Paired items
\end{minipage} & \begin{minipage}[b]{\linewidth}\raggedright
C
\end{minipage} & \begin{minipage}[b]{\linewidth}\raggedright
F
\end{minipage} & \begin{minipage}[b]{\linewidth}\raggedright
C\ensuremath{-}F
\end{minipage} & \begin{minipage}[b]{\linewidth}\raggedright
Nominal 95\% interval
\end{minipage} \\
\midrule\noalign{}
\endfirsthead
\toprule\noalign{}
\begin{minipage}[b]{\linewidth}\raggedright
Criterion
\end{minipage} & \begin{minipage}[b]{\linewidth}\raggedright
Paired items
\end{minipage} & \begin{minipage}[b]{\linewidth}\raggedright
C
\end{minipage} & \begin{minipage}[b]{\linewidth}\raggedright
F
\end{minipage} & \begin{minipage}[b]{\linewidth}\raggedright
C\ensuremath{-}F
\end{minipage} & \begin{minipage}[b]{\linewidth}\raggedright
Nominal 95\% interval
\end{minipage} \\
\midrule\noalign{}
\endhead
\bottomrule\noalign{}
\endlastfoot
Correctness, all questions & 100 & 1.360 & 1.370 & \ensuremath{-}0.010 & {[}\ensuremath{-}0.157, +0.130{]} \\
Completeness, answerable & 90 & 1.570 & 1.344 & +0.226 & {[}+0.115, +0.344{]} \\
Evidence support (gold passages) & 90 & 1.456 & 1.422 & +0.033 & {[}\ensuremath{-}0.085, +0.156{]} \\
Temporal accuracy, numeric pairs & 59 & 1.887 & 1.831 & +0.056 & {[}\ensuremath{-}0.017, +0.136{]} \\
Readability, numeric pairs & 99 & 1.806 & 1.680 & +0.126 & {[}\ensuremath{-}0.002, +0.261{]} \\
\end{longtable}

Each question contributes the average of its available matched-run scores. Correctness includes 99 questions with three paired runs and one with two. Temporal accuracy includes 56, 1 and 2 questions with three, two and one matched runs; readability includes 97 and 2 questions with three and two. Completeness and evidence support include all 90 answerable questions over three runs. Evidence support is judged against common gold passages, not each system's retrieved context.

Completeness survives the protocol change; correctness does not show an advantage under either protocol. The answer-only readability estimate favors C slightly, which does not reproduce the human preference for F (Section 7.2). Other configuration contrasts, including J\ensuremath{-}C, have not been re-graded under the answer-only protocol.

\subsection{Scoring population}

Which questions enter the mean reverses some contrasts. On answerable questions, A and B score about 0.10 above F (Table 7). Over all 100 questions, both fall slightly below it: A\ensuremath{-}F = \ensuremath{-}0.037 {[}\ensuremath{-}0.213, +0.140{]} and B\ensuremath{-}F = \ensuremath{-}0.010 {[}\ensuremath{-}0.170, +0.157{]} (Table 8). The reason is F's behavior on the ten unanswerable items, where its clean refusals earn full credit.

The same effect shrinks C's advantage. Over the 90 answerable questions, the paired contrasts are C\ensuremath{-}F = +0.141 {[}+0.015, +0.274{]}, A\ensuremath{-}F = +0.100 {[}\ensuremath{-}0.063, +0.267{]} and B\ensuremath{-}F = +0.093 {[}\ensuremath{-}0.063, +0.256{]}. Only C\ensuremath{-}F excludes zero, and only on this population; over all 100 questions it falls to +0.060 {[}\ensuremath{-}0.080, +0.200{]}. With ten unanswerable items scored 0--2, they can move a 100-question mean by up to 0.2 points, more than any internal contrast in Table 8.

\subsection{Abstention coding}

The two systems fail in opposite directions, and a binary abstention metric hides both. C never cleanly refuses: on unanswerable questions it mostly hedges, acknowledging the gap and then supplying adjacent information. F cleanly refuses every unanswerable question but also refuses 18--19 of 90 answerable questions per run (Table 11).

\begin{longtable}[]{@{}lll@{}}
\caption{Rule-coded refusal / hedge / answer counts, Qwen runs 1, 2 and 3.}\tabularnewline
\toprule\noalign{}
Config & Unanswerable (10 per run) & Answerable (90 per run) \\
\midrule\noalign{}
\endfirsthead
\toprule\noalign{}
Config & Unanswerable (10 per run) & Answerable (90 per run) \\
\midrule\noalign{}
\endhead
\bottomrule\noalign{}
\endlastfoot
C & 0/9/1; 0/8/2; 0/7/3 & 0/1/89; 0/0/90; 0/1/89 \\
F & 10/0/0; 10/0/0; 10/0/0 & 18/0/72; 19/0/71; 18/0/72 \\
G & 7/3/0; 5/5/0; 6/4/0 & 11/1/78; 9/6/75; 7/7/76 \\
H & 0/7/3; 0/8/2; 0/7/3 & 0/2/88; 0/1/89; 0/1/89 \\
I & 8/2/0; 8/2/0; 8/2/0 & 7/5/78; 8/4/78; 6/4/80 \\
\end{longtable}

With ten unanswerable items per run, refusal rates are imprecise. In run 1, F's clean-refusal rate on unanswerable items is 10/10 (95\% Wilson interval 0.72--1.00) and C's is 0/10 (0.00--0.28). C hedged on 9/10 (0.60--0.98). F's false-refusal rate on answerable items is 18/90 (0.13--0.29). The three runs reuse the same questions, so we report per-run intervals rather than pooling.

The guarded variants G and I sit between the two: more clean refusals on unanswerable items than C, and fewer false refusals than F (I: 6--8 of 90 per run). No configuration dominates on every dimension. Describing F as simply more reliable would omit its false-refusal burden; describing C as answering everything would omit that most of its unanswerable responses are hedges.

A binary metric scores these behaviors inconsistently. If a hedge counts as abstention, C appears to abstain correctly; if it counts as an answer, C appears to answer unsupported questions confidently. Neither describes what the reader sees. The judge's own abstention field agreed with human refusal verdicts on only 45 of 80 run-1 unanswerable cases, while the rule-based coding agreed on 67 of 80 (\ensuremath{\kappa} = 0.675) (Section 7.3).

In 8 of 18, 9 of 19 and 8 of 18 of F's false refusals per run, at least one gold passage appeared among F's eight returned sources; in one case per run, all gold passages did. About half of F's false refusals therefore occur despite retrieved supporting evidence, which implicates its strict refusal instruction; the rest follow retrieval misses.

\subsection{Retrieval normalization}

How retrieval is counted changes the explanation for LightRAG's deficit. By source-count recall@5, J recovers 0.147 of annotated evidence against C's 0.481. By the uncapped all-source ceiling, J recovers more than C (0.594 against 0.560). J's problem is evidence density in a long source list, not missing evidence (Table 12).

\begin{longtable}[]{@{}
  >{\raggedright\arraybackslash}p{(\linewidth - 14\tabcolsep) * \real{0.1250}}
  >{\raggedright\arraybackslash}p{(\linewidth - 14\tabcolsep) * \real{0.1250}}
  >{\raggedright\arraybackslash}p{(\linewidth - 14\tabcolsep) * \real{0.1250}}
  >{\raggedright\arraybackslash}p{(\linewidth - 14\tabcolsep) * \real{0.1250}}
  >{\raggedright\arraybackslash}p{(\linewidth - 14\tabcolsep) * \real{0.1250}}
  >{\raggedright\arraybackslash}p{(\linewidth - 14\tabcolsep) * \real{0.1250}}
  >{\raggedright\arraybackslash}p{(\linewidth - 14\tabcolsep) * \real{0.1250}}
  >{\raggedright\arraybackslash}p{(\linewidth - 14\tabcolsep) * \real{0.1250}}@{}}
\caption{Gold-evidence recall over returned sources, averaged over three Qwen runs (main budget: 3,000 characters).}\tabularnewline
\toprule\noalign{}
\begin{minipage}[b]{\linewidth}\raggedright
Config
\end{minipage} & \begin{minipage}[b]{\linewidth}\raggedright
3k chars
\end{minipage} & \begin{minipage}[b]{\linewidth}\raggedright
2k chars
\end{minipage} & \begin{minipage}[b]{\linewidth}\raggedright
5k chars
\end{minipage} & \begin{minipage}[b]{\linewidth}\raggedright
10k chars
\end{minipage} & \begin{minipage}[b]{\linewidth}\raggedright
All-source ceiling
\end{minipage} & \begin{minipage}[b]{\linewidth}\raggedright
Recall@5
\end{minipage} & \begin{minipage}[b]{\linewidth}\raggedright
Recall@8
\end{minipage} \\
\midrule\noalign{}
\endfirsthead
\toprule\noalign{}
\begin{minipage}[b]{\linewidth}\raggedright
Config
\end{minipage} & \begin{minipage}[b]{\linewidth}\raggedright
3k chars
\end{minipage} & \begin{minipage}[b]{\linewidth}\raggedright
2k chars
\end{minipage} & \begin{minipage}[b]{\linewidth}\raggedright
5k chars
\end{minipage} & \begin{minipage}[b]{\linewidth}\raggedright
10k chars
\end{minipage} & \begin{minipage}[b]{\linewidth}\raggedright
All-source ceiling
\end{minipage} & \begin{minipage}[b]{\linewidth}\raggedright
Recall@5
\end{minipage} & \begin{minipage}[b]{\linewidth}\raggedright
Recall@8
\end{minipage} \\
\midrule\noalign{}
\endhead
\bottomrule\noalign{}
\endlastfoot
A & 0.4824 & 0.4667 & 0.5009 & 0.5009 & 0.5009 & 0.4630 & 0.5009 \\
B & 0.4750 & 0.4593 & 0.4935 & 0.4935 & 0.4935 & 0.4556 & 0.4935 \\
C & 0.5170 & 0.5016 & 0.5472 & 0.5596 & 0.5596 & 0.4806 & 0.5386 \\
D & 0.4787 & 0.4630 & 0.4972 & 0.4972 & 0.4972 & 0.4593 & 0.4972 \\
F & 0.5648 & 0.5315 & 0.5981 & 0.6204 & 0.6204 & 0.5648 & 0.6204 \\
G & 0.4972 & 0.4837 & 0.5237 & 0.5337 & 0.5337 & 0.4623 & 0.5124 \\
H & 0.4997 & 0.4840 & 0.5275 & 0.5312 & 0.5312 & 0.4661 & 0.5163 \\
I & 0.4938 & 0.4707 & 0.5259 & 0.5309 & 0.5309 & 0.4534 & 0.5071 \\
J & 0.2907 & 0.2130 & 0.3926 & 0.4975 & 0.5938 & 0.1472 & 0.1975 \\
\end{longtable}

All cells use the same 90 answerable questions per run and are independent of judge success. E shares C's sources and is omitted. Recall@5 and recall@8 count different source units across systems and are secondary.

F's plain cosine top-k exceeds every internal configuration at every budget and in every run. Planning and multi-task retrieval lift C above the single-task paths A, B and D (0.475--0.482 at 3k), but not to F's level. Across runs, J returns 19,181.9 source characters on answerable questions against C's 3,392.7. In run-1 usage records, J's final generation call has a median of 17,343.5 prompt tokens against 3,138.5 for C's synthesizer. Access to more prompt text did not remove J's correctness deficit.

C recovers less annotated evidence than F at every budget, yet gives more complete answers (Table 10). Its extra content may come from passages outside the gold annotations, from candidate drafts, or from the generator's own knowledge. To locate the source of C's additional completeness, we audited 20 C answers drawn by stratified random sampling (seed 20261020) from the 62 answerable run--question pairs in which C's answer-only completeness exceeded F's. The audit was a single-coder check performed with language-model assistance. Each answer was split into atomic factual claims (n = 195), which were checked against the passage text the synthesizer received (1,400 characters per passage) and against the intermediate drafts. Of these claims, 24\% were supported by an annotated gold passage and 55\% by a non-gold returned passage, usually an unannotated passage of the same report; 5\% appeared only in an intermediate draft, 3\% in no returned source, and 10\% contradicted a returned source, mostly through misread table cells and misattributed figures. C's advantage rested on gold content that F did not state (10 cases, five of them F refusals) or on non-gold passages (9 cases), which suggests that the gold annotations under-count the evidence C uses rather than that C relies on parametric knowledge. However, 14 of the 20 answers contained at least one draft-only, unsupported or contradicted claim. To check the labels, GPT-5 (gpt-5-2025-08-07, reasoning effort high) re-coded the same 195 claims blind to the original labels and notes, given the same case material, one request per answer. It agreed on 178 claims (91\%; Cohen's \ensuremath{\kappa} = 0.86, 95\% bootstrap interval over answers 0.79--0.93); six of the 17 disagreements were claims the original coder had left unclear. Under the re-code the shares are 25\% gold, 55\% non-gold, 6\% draft-only, 4\% unsourced and 11\% contradicted, and 16 of the 20 answers contain at least one such claim. The claim segmentation is the original coder's, and agreement with one language model is not human validation. These curves measure emitted source lists, not the exact generation context, and do not explain correctness differences causally.

\subsection{Inference cost}

C used 3.45 times F's recorded model calls in Qwen run 1 and 3.40 times with EXAONE. Its prompt tokens were 3.6 times F's (Table 13). These ratios are lower bounds: for every configuration except F and J, the recorded counts omit the backend candidate-drafting call issued for each retrieval call (150 for C in run 1, raising its total to about 495), and that call's tokens were not recorded.

\begin{longtable}[]{@{}llll@{}}
\caption{Online inference resources per 100 questions, Qwen run 1.}\tabularnewline
\toprule\noalign{}
Config & LLM calls & Prompt tokens & Completion tokens \\
\midrule\noalign{}
\endfirsthead
\toprule\noalign{}
Config & LLM calls & Prompt tokens & Completion tokens \\
\midrule\noalign{}
\endhead
\bottomrule\noalign{}
\endlastfoot
A & 100 & 279,036 & 61,053 \\
B & 100 & 281,264 & 62,474 \\
C & 345 & 910,186 & 173,459 \\
D & 200 & 664,312 & 146,913 \\
E (derived) & 245 & 398,997 & 79,080 \\
F & 100 & 254,375 & 17,048 \\
G & 345 & 888,683 & 148,781 \\
H & 347 & 894,599 & 171,586 \\
I & 347 & 888,754 & 153,895 \\
J & 200 & 1,805,483 & 49,588 \\
\end{longtable}

A and B record the same number of calls as F (each also issues one unrecorded backend drafting call) and score within C's run-to-run range on answerable questions, using less than a third of C's prompt tokens. E's derived record isolates the diagnostic verifier's accounting overhead (100 calls) without regeneration. J uses two calls per question, including keyword extraction, and the largest prompt-token footprint. Call counts are not monetary cost, because requests differ in length; offline indexing is excluded. Wall-clock times were recorded on a shared server under uncontrolled load and are reported only in Supplement S1.

\subsection{Error analysis}

With three runs, each item difference \(d_i\) (Section 5.6) is a multiple of 1/3. Sixty-two of the 100 questions have \(d_i\) = 0 exactly, so the C\ensuremath{-}F estimate of +0.060 rests on the remaining 38 (23 favor C, 15 favor F; \(\Sigma d_i\) = +6.0). A threshold of \(|d_i|\) \ensuremath{\geq} 2/3 would have selected only 28 questions, so we analysed all 38. For each, the coder read the question, reference answer, required points, both systems' answers from all three runs and the judge's explanations, and assigned the losing system one primary error category and, where one clearly applied, a secondary category. Coding was done by a single coder with language-model assistance; it was not blind to configuration, and the judge's reasoning was visible. 

To check this coding, the 38 questions were re-coded blind by GPT-5 (gpt-5-2025-08-07, reasoning effort high; the model accepts no temperature setting). It received the question, reference answer, required points, gold passages and both answers from all three runs with their judge scores, but no configuration names, judge explanations or citation markers, and the order of the two answers was randomized per question. It used the same category definitions and instructions, one request per question, with output restricted to the category names. The re-code identified the same losing system in all 38 questions and the same primary category in 26 (68\%; Cohen's \ensuremath{\kappa} = 0.63, 95\% item-bootstrap interval 0.45--0.78). Most disagreements lie at two boundaries: three missing-hop codes became incomplete answers, and seven wrong-period, adjacent-fact, incomplete or other codes became wrong region or indicator, the category GPT-5 used most (13 questions against 6). Both codings attribute every unjustified refusal to F and every adjacent-fact substitution and unsupported addition to C, but GPT-5 assigns more of C's losses to wrong region or indicator (5 against 1; Table 14). Agreement with one language model supports the reproducibility of these categories; it is not human validation.

The two systems lose in different ways (Table 14). F's losses are mainly omissions: six clean refusals of answerable questions, which carry the largest share of the absolute difference (10.7 of 36.7), and six incomplete answers, two of them partial refusals. F also makes seven errors of reading the wrong period, row or indicator, mostly from tables it had retrieved. C never loses by refusing. Its losses are mainly additions: five answers whose details appear in no returned source, and five adjacent-fact substitutions. Four of those substitutions are hedges on unanswerable questions, which F won with a clean refusal in every run; in one run a hedge became a confident false answer (cul\_unans\_01). Retrieval was implicated less often in C's losses than in F's: on answerable items, the losing system's returned sources held less gold evidence than the winner's in 3 of C's 11 losses and 10 of F's 23. In 9 of the 38 questions the winning system made the same error in at least one run, so many differences are of frequency rather than kind. Across all 100 questions, F's correctness was identical in all three runs on 97 questions and C's on 79, so run-to-run movement in C\ensuremath{-}F comes almost entirely from C.

\begin{longtable}[]{@{}
  >{\raggedright\arraybackslash}p{(\linewidth - 8\tabcolsep) * \real{0.4444}}
  >{\raggedright\arraybackslash}p{(\linewidth - 8\tabcolsep) * \real{0.1111}}
  >{\raggedright\arraybackslash}p{(\linewidth - 8\tabcolsep) * \real{0.1111}}
  >{\raggedright\arraybackslash}p{(\linewidth - 8\tabcolsep) * \real{0.1111}}
  >{\raggedright\arraybackslash}p{(\linewidth - 8\tabcolsep) * \real{0.2222}}@{}}
\caption{Primary error of the losing system on the 38 questions with non-zero item-averaged C\ensuremath{-}F correctness (Qwen, trace-aware grading; single model-assisted coder; the last column gives the blinded GPT-5 re-code as C / F losses).}\tabularnewline
\toprule\noalign{}
\begin{minipage}[b]{\linewidth}\raggedright
Primary error
\end{minipage} & \begin{minipage}[b]{\linewidth}\raggedright
C loses
\end{minipage} & \begin{minipage}[b]{\linewidth}\raggedright
F loses
\end{minipage} & \begin{minipage}[b]{\linewidth}\raggedright
\(\Sigma \lvert d_i\rvert\)
\end{minipage} & \begin{minipage}[b]{\linewidth}\raggedright
GPT-5 re-code (C / F)
\end{minipage} \\
\midrule\noalign{}
\endfirsthead
\toprule\noalign{}
\begin{minipage}[b]{\linewidth}\raggedright
Primary error
\end{minipage} & \begin{minipage}[b]{\linewidth}\raggedright
C loses
\end{minipage} & \begin{minipage}[b]{\linewidth}\raggedright
F loses
\end{minipage} & \begin{minipage}[b]{\linewidth}\raggedright
\(\Sigma \lvert d_i\rvert\)
\end{minipage} & \begin{minipage}[b]{\linewidth}\raggedright
GPT-5 re-code (C / F)
\end{minipage} \\
\midrule\noalign{}
\endhead
\bottomrule\noalign{}
\endlastfoot
Unjustified refusal (answerable item) & 0 & 6 & 10.67 & 0 / 7 \\
Adjacent-fact substitution & 5 & 0 & 7.67 & 3 / 0 \\
Wrong region or indicator & 1 & 5 & 5.33 & 5 / 8 \\
Incomplete (missing required points) & 2 & 6 & 4.67 & 1 / 7 \\
Unsupported added content & 5 & 0 & 3.33 & 5 / 0 \\
Missing hop & 1 & 2 & 2.33 & 0 / 0 \\
Wrong period & 0 & 2 & 1.00 & 0 / 1 \\
Other & 1 & 2 & 1.67 & 0 / 0 \\
Judge inconsistency (answer acceptable) & 0 & 0 & 0.00 & 1 / 0 \\
Total & 15 & 23 & 36.67 & 15 / 23 \\
\end{longtable}

By question type (C losses / F losses; net \(\Sigma d_i\)): factual 3/5 (+1.67); comparison 3/9 (+4.67); multi-hop 5/9 (+6.33); unanswerable 4/0 (\ensuremath{-}6.67). In the original coding, judge inconsistency was never the primary category, but it applied as a secondary category in three questions where near-identical answers received different scores across runs.

\emph{Example 1, unjustified refusal (reg\_comp\_03, d = +1.67).} F answered ``검색된 자료에서 확인할 수 없습니다'' (``This cannot be confirmed from the retrieved material'') in all three runs, although both gold passages were in its source list. C reported the longest commute times (Eumseong 57.2 and Jincheon 56.3 minutes) against the cities' 40.7--48.9 minutes.

\emph{Example 2, hedge with adjacent-fact substitution (reg\_unans\_01, d = \ensuremath{-}2.0).} Asked for Cheongju's 2024 budget for its demand-responsive bus service, C wrote: ``구체적인 수치는 제공된 자료에 직접적으로 명시되어 있지 않습니다. 그러나 청주시가 2019년과 2020년에 시내(농어촌)버스 운영에 투입한 재정지원 규모를 살펴보면\ldots{}'' (``The specific figure is not stated in the provided material. However, looking at the subsidies Cheongju gave city and rural bus operations in 2019 and 2020\ldots{}''). It then gave general bus subsidies from another period and service. F refused cleanly.

\emph{Example 3, unsupported added content (env\_hop\_02, d = \ensuremath{-}1.0).} C named ``한국도로공사 단양시멘트공장과 대한제강 단양공장'' (``the Korea Expressway Corporation Danyang cement plant and the Daehan Steel Danyang plant'') as Danyang's cement plants; the source report lists Sungshin, Hanil and Hanil Hyundai. Neither name appears in any returned source in any run. Both first appear in the retriever stage's candidate draft and pass through the synthesizer unchanged.

\subsection{Protocol sensitivity across all contrasts}

Applying the profile of Section 5.6 to all eleven paired contrasts shows that protocol dependence is not specific to C\ensuremath{-}F (Table 15). No correctness verdict is protocol-robust. Seven of the eleven correctness contrasts take both signs across protocols, and the ranges of the point estimates run from 0.101 (J\ensuremath{-}F) to 0.354 (C\ensuremath{-}F). C\ensuremath{-}F has the widest range because it is the only contrast measured under both generators and both judge packets. Its interval excludes zero in 4 of its 24 protocols, all of them Qwen protocols scored on the 90 answerable questions. J\ensuremath{-}C is the most stable correctness contrast: negative under every protocol, with an interval below zero in five of eight.

Completeness behaves differently. Four contrasts are protocol-robust: C\ensuremath{-}F (12 of 12 protocols), A\ensuremath{-}F and B\ensuremath{-}F (4 of 4 each) and J\ensuremath{-}C (4 of 4, favoring C). The robust completeness advantage over vanilla RAG is therefore shared by the simplified paths A and B, which make one retrieval call without a planner, selector or verifier. Their estimates (+0.222 to +0.233) lie within C's range (+0.167 to +0.311), and the C\ensuremath{-}A and C\ensuremath{-}B completeness intervals include zero under every protocol. Within this study, the completeness gain over F is associated with the retrieval, drafting and synthesis components that A, B and C share, not with the planner, selector and verifier that C adds. G and I trail C on completeness under every protocol, with intervals below zero in two of four; H's completeness estimate changes sign (\ensuremath{-}0.100 to +0.033).

\begin{longtable}[]{@{}
  >{\raggedright\arraybackslash}p{(\linewidth - 14\tabcolsep) * \real{0.0870}}
  >{\raggedright\arraybackslash}p{(\linewidth - 14\tabcolsep) * \real{0.1739}}
  >{\raggedright\arraybackslash}p{(\linewidth - 14\tabcolsep) * \real{0.1304}}
  >{\raggedright\arraybackslash}p{(\linewidth - 14\tabcolsep) * \real{0.2174}}
  >{\raggedright\arraybackslash}p{(\linewidth - 14\tabcolsep) * \real{0.0870}}
  >{\raggedright\arraybackslash}p{(\linewidth - 14\tabcolsep) * \real{0.0870}}
  >{\raggedright\arraybackslash}p{(\linewidth - 14\tabcolsep) * \real{0.1304}}
  >{\raggedright\arraybackslash}p{(\linewidth - 14\tabcolsep) * \real{0.0870}}@{}}
\caption{Protocol-sensitivity profile of every paired contrast. \ensuremath{\Delta} range is the minimum to maximum point estimate over the protocols. Decisions count the protocols whose 95\% interval lies above zero, includes zero or lies below zero. Correctness protocols cross the run set (runs 1, 2 and 3, and the three-run pool) with the scoring population (100 questions or the 90 answerable ones); C\ensuremath{-}F also crosses generator and judge packet. Completeness uses the 90 answerable questions.}\tabularnewline
\toprule\noalign{}
\begin{minipage}[b]{\linewidth}\raggedright
Contrast
\end{minipage} & \begin{minipage}[b]{\linewidth}\raggedright
Criterion
\end{minipage} & \begin{minipage}[b]{\linewidth}\raggedright
Protocols
\end{minipage} & \begin{minipage}[b]{\linewidth}\raggedright
\ensuremath{\Delta} range
\end{minipage} & \begin{minipage}[b]{\linewidth}\raggedright
Width
\end{minipage} & \begin{minipage}[b]{\linewidth}\raggedright
Sign change
\end{minipage} & \begin{minipage}[b]{\linewidth}\raggedright
Decisions + / 0 / \ensuremath{-}
\end{minipage} & \begin{minipage}[b]{\linewidth}\raggedright
Robust
\end{minipage} \\
\midrule\noalign{}
\endfirsthead
\toprule\noalign{}
\begin{minipage}[b]{\linewidth}\raggedright
Contrast
\end{minipage} & \begin{minipage}[b]{\linewidth}\raggedright
Criterion
\end{minipage} & \begin{minipage}[b]{\linewidth}\raggedright
Protocols
\end{minipage} & \begin{minipage}[b]{\linewidth}\raggedright
\ensuremath{\Delta} range
\end{minipage} & \begin{minipage}[b]{\linewidth}\raggedright
Width
\end{minipage} & \begin{minipage}[b]{\linewidth}\raggedright
Sign change
\end{minipage} & \begin{minipage}[b]{\linewidth}\raggedright
Decisions + / 0 / \ensuremath{-}
\end{minipage} & \begin{minipage}[b]{\linewidth}\raggedright
Robust
\end{minipage} \\
\midrule\noalign{}
\endhead
\bottomrule\noalign{}
\endlastfoot
C\ensuremath{-}F & Correctness & 24 & \ensuremath{-}0.121 to +0.233 & 0.354 & Yes & 4 / 20 / 0 & No \\
C\ensuremath{-}F & Completeness & 12 & +0.167 to +0.311 & 0.144 & No & 12 / 0 / 0 & Yes \\
C\ensuremath{-}A & Correctness & 8 & \ensuremath{-}0.022 to +0.140 & 0.162 & Yes & 1 / 7 / 0 & No \\
C\ensuremath{-}A & Completeness & 4 & +0.000 to +0.078 & 0.078 & No & 0 / 4 / 0 & No \\
C\ensuremath{-}B & Correctness & 8 & +0.000 to +0.190 & 0.190 & No & 2 / 6 / 0 & No \\
C\ensuremath{-}B & Completeness & 4 & +0.000 to +0.089 & 0.089 & No & 0 / 4 / 0 & No \\
C\ensuremath{-}D & Correctness & 8 & +0.000 to +0.140 & 0.140 & No & 2 / 6 / 0 & No \\
C\ensuremath{-}D & Completeness & 4 & +0.011 to +0.111 & 0.100 & No & 1 / 3 / 0 & No \\
A\ensuremath{-}F & Correctness & 8 & \ensuremath{-}0.080 to +0.122 & 0.202 & Yes & 0 / 8 / 0 & No \\
A\ensuremath{-}F & Completeness & 4 & +0.222 to +0.233 & 0.011 & No & 4 / 0 / 0 & Yes \\
B\ensuremath{-}F & Correctness & 8 & \ensuremath{-}0.040 to +0.100 & 0.140 & Yes & 0 / 8 / 0 & No \\
B\ensuremath{-}F & Completeness & 4 & +0.222 to +0.233 & 0.011 & No & 4 / 0 / 0 & Yes \\
G\ensuremath{-}C & Correctness & 8 & \ensuremath{-}0.189 to +0.030 & 0.219 & Yes & 0 / 6 / 2 & No \\
G\ensuremath{-}C & Completeness & 4 & \ensuremath{-}0.111 to \ensuremath{-}0.044 & 0.067 & No & 0 / 2 / 2 & No \\
H\ensuremath{-}C & Correctness & 8 & \ensuremath{-}0.150 to +0.056 & 0.206 & Yes & 0 / 6 / 2 & No \\
H\ensuremath{-}C & Completeness & 4 & \ensuremath{-}0.100 to +0.033 & 0.133 & Yes & 0 / 3 / 1 & No \\
I\ensuremath{-}C & Correctness & 8 & \ensuremath{-}0.178 to +0.080 & 0.258 & Yes & 0 / 7 / 1 & No \\
I\ensuremath{-}C & Completeness & 4 & \ensuremath{-}0.133 to \ensuremath{-}0.011 & 0.122 & No & 0 / 2 / 2 & No \\
J\ensuremath{-}C & Correctness & 8 & \ensuremath{-}0.340 to \ensuremath{-}0.172 & 0.168 & No & 0 / 3 / 5 & No \\
J\ensuremath{-}C & Completeness & 4 & \ensuremath{-}0.300 to \ensuremath{-}0.156 & 0.144 & No & 0 / 0 / 4 & Yes \\
J\ensuremath{-}F & Correctness & 8 & \ensuremath{-}0.190 to \ensuremath{-}0.089 & 0.101 & No & 0 / 7 / 1 & No \\
J\ensuremath{-}F & Completeness & 4 & +0.011 to +0.067 & 0.056 & No & 0 / 4 / 0 & No \\
\end{longtable}

Single-run protocols are noisier than pooled ones by construction, so part of each range reflects run-to-run variation (Section 6.1). Restricted to the three-run pooled protocols, one correctness verdict becomes robust: J\ensuremath{-}C, whose intervals lie below zero over both scoring populations. C\ensuremath{-}F correctness does not: its interval excludes zero in only one of six pooled protocols (Qwen, trace-aware grading, answerable questions). The four completeness verdicts that are robust over all protocols remain robust, and the pooled completeness intervals of G\ensuremath{-}C and I\ensuremath{-}C also lie below zero; for these and the other non-C\ensuremath{-}F contrasts, the pooled family has a single completeness protocol.

\section{Human validation}

Blinded human ratings confirm the pipeline's advantage in completeness and temporal accuracy, favor vanilla RAG on readability, and show no reliable correctness difference. Ratings cover run-1 Qwen C and F answers only.

\subsection{Design}

Six reviewers rated paired C/F answers. Five each reviewed their assigned policy domain plus a shared ten-question block (five answerable, five unanswerable), for 28 cards each; a sixth (R6) reviewed all 100 questions. Cards showed the question, reference answer, gold passages and two answers under randomized labels, with system identities and source counts hidden. Both answers had to be locked before a card counted as complete. The final data comprise 240 cards and 480 answer ratings, with numeric correctness on all 240 pairs; every question has at least one paired correctness rating.

For each criterion, C\ensuremath{-}F is computed within reviewer and question, averaged across reviewers within question, then across questions. Intervals condition on this reviewer panel and do not represent the wider population of experts.

\textbf{Protocol deviation.} R6's ten unanswerable cards were re-rated once under a clarified instruction. In the first pass, R6 had used cannot\_assess for correctness on eight of them, and had read the refusal item as whether refusal would be appropriate rather than whether the answer refused. During re-rating, an interface defect overwrote newly entered scores on re-opened answers with their previous values; it was fixed and the affected answers re-entered. First-time ratings were not affected, every unlock is logged with the prior rating file, and no score was entered or changed by the authors. With R6's first-pass ratings, excluding the eight cannot\_assess cards, human correctness C\ensuremath{-}F is +0.142 {[}+0.004, +0.283{]}; with the re-rated cards it is +0.117 {[}\ensuremath{-}0.023, +0.258{]} (Table 16). The re-rating moves the lower bound across zero, so the human correctness result is borderline and depends on how R6's unanswerable cards are treated.

\subsection{Answer-quality results}

\begin{longtable}[]{@{}
  >{\raggedright\arraybackslash}p{(\linewidth - 6\tabcolsep) * \real{0.2500}}
  >{\raggedright\arraybackslash}p{(\linewidth - 6\tabcolsep) * \real{0.2500}}
  >{\raggedright\arraybackslash}p{(\linewidth - 6\tabcolsep) * \real{0.2500}}
  >{\raggedright\arraybackslash}p{(\linewidth - 6\tabcolsep) * \real{0.2500}}@{}}
\caption{Paired human answer-quality contrasts, C minus F.}\tabularnewline
\toprule\noalign{}
\begin{minipage}[b]{\linewidth}\raggedright
Criterion
\end{minipage} & \begin{minipage}[b]{\linewidth}\raggedright
Questions with paired scores
\end{minipage} & \begin{minipage}[b]{\linewidth}\raggedright
C\ensuremath{-}F
\end{minipage} & \begin{minipage}[b]{\linewidth}\raggedright
Nominal 95\% interval
\end{minipage} \\
\midrule\noalign{}
\endfirsthead
\toprule\noalign{}
\begin{minipage}[b]{\linewidth}\raggedright
Criterion
\end{minipage} & \begin{minipage}[b]{\linewidth}\raggedright
Questions with paired scores
\end{minipage} & \begin{minipage}[b]{\linewidth}\raggedright
C\ensuremath{-}F
\end{minipage} & \begin{minipage}[b]{\linewidth}\raggedright
Nominal 95\% interval
\end{minipage} \\
\midrule\noalign{}
\endhead
\bottomrule\noalign{}
\endlastfoot
Correctness & 100 & +0.117 & {[}\ensuremath{-}0.023, +0.258{]} \\
Completeness & 90 & +0.367 & {[}+0.250, +0.491{]} \\
Evidence support & 90 & +0.063 & {[}\ensuremath{-}0.074, +0.209{]} \\
Temporal accuracy & 90 & +0.294 & {[}+0.173, +0.425{]} \\
Readability & 64 & \ensuremath{-}0.313 & {[}\ensuremath{-}0.477, \ensuremath{-}0.146{]} \\
\end{longtable}

Reviewers favor C on completeness and temporal accuracy, and F on readability. The correctness estimate is positive but its interval includes zero, in line with both machine protocols. Readability was optional and sparse for several reviewers, so its result applies to the 64 questions with paired ratings. The five intervals are not simultaneous.

The temporal-accuracy advantage is directly relevant to the application: policy questions require the right reporting period. Machine answer-only grading points the same way but more weakly (+0.056 {[}\ensuremath{-}0.017, +0.136{]}, Table 10).

\subsection{Reviewer sensitivity and agreement}

Reviewers split in direction: four favor C on correctness and two favor F (Table 17). All six are retained; a preference or a high scoring level is not an exclusion criterion.

\begin{longtable}[]{@{}llllll@{}}
\caption{Correctness by reviewer.}\tabularnewline
\toprule\noalign{}
Reviewer & Assigned domain & Numeric pairs & C mean & F mean & C\ensuremath{-}F \\
\midrule\noalign{}
\endfirsthead
\toprule\noalign{}
Reviewer & Assigned domain & Numeric pairs & C mean & F mean & C\ensuremath{-}F \\
\midrule\noalign{}
\endhead
\bottomrule\noalign{}
\endlastfoot
R1 & Industry & 28 & 1.536 & 1.393 & +0.143 \\
R2 & Region & 28 & 1.393 & 1.214 & +0.179 \\
R3 & Environment & 28 & 1.286 & 1.107 & +0.179 \\
R4 & Population & 28 & 1.214 & 1.643 & \ensuremath{-}0.429 \\
R5 & Culture & 28 & 0.821 & 1.036 & \ensuremath{-}0.214 \\
R6 & All domains & 100 & 1.710 & 1.570 & +0.140 \\
\end{longtable}

\begin{longtable}[]{@{}lll@{}}
\caption{Human correctness C\ensuremath{-}F, leaving out one reviewer at a time.}\tabularnewline
\toprule\noalign{}
Reviewer left out & C\ensuremath{-}F & Nominal 95\% interval \\
\midrule\noalign{}
\endfirsthead
\toprule\noalign{}
Reviewer left out & C\ensuremath{-}F & Nominal 95\% interval \\
\midrule\noalign{}
\endhead
\bottomrule\noalign{}
\endlastfoot
R1 & +0.120 & {[}\ensuremath{-}0.027, +0.269{]} \\
R2 & +0.117 & {[}\ensuremath{-}0.029, +0.265{]} \\
R3 & +0.102 & {[}\ensuremath{-}0.033, +0.241{]} \\
R4 & +0.167 & {[}+0.035, +0.303{]} \\
R5 & +0.154 & {[}+0.010, +0.300{]} \\
R6 & +0.040 & {[}\ensuremath{-}0.128, +0.206{]} \\
\end{longtable}

The interval includes zero in four of six analyses and excludes it only when a reviewer who favors F is removed. Removing R6 moves the estimate most, for structural reasons. R6 is one of two reviewers on each of the 90 questions outside the shared block, so R6 accounts on average for 0.467 of each question's paired difference. R6 also rates near the top of the scale (80\% of C and 73\% of F correctness ratings are 2), and 70 of R6's 100 pairs are ties; the remaining 30 split 21 for C and 9 for F. The no-reliable-difference conclusion holds with and without R6.

Three of R6's re-entered C correctness scores on unanswerable questions (2, 2 and 1) conflict with R6's own verdict that those answers did not refuse correctly, and with the rubric for substituted figures. Scoring them as 0 gives +0.092 {[}\ensuremath{-}0.050, +0.238{]}. The ratings are reported as recorded.

\textbf{Agreement.} Pooled judge--human correctness agreement is 65.6\%, with quadratic-weighted \ensuremath{\kappa} = 0.520 over 480 paired observations. By criterion, \ensuremath{\kappa} is 0.650 for completeness, 0.531 for evidence support, 0.627 for temporal accuracy and 0.322 for readability. Pooled human--human correctness agreement is 50.8\% (weighted \ensuremath{\kappa} = 0.354) over 480 pairwise observations, using every question a reviewer pair shares. These observations reuse answers and are not independent; they are descriptive, not a six-rater reliability estimate. Agreement statistics concern the trace-aware grades; human agreement with the answer-only grades has not been computed.

On run-1 unanswerable questions, 67 of 80 human refusal verdicts agree with the rule-based coding's clean-refusal decision (Cohen's \ensuremath{\kappa} = 0.675) (C never cleanly refuses; F always does), and 8 of the 13 disagreements come from one reviewer. The judge's abstention field agrees on only 45 of 80. Specialist differences outside the shared block combine reviewer and domain effects.

\section{Discussion}

For a policy researcher, MARO's answers are fuller, better dated and checkable. Blinded reviewers favored its completeness by +0.367 and its temporal accuracy by +0.294, answer-only grading confirmed the completeness gain, and every sentence cites a numbered passage with the verifier's claim list beside the answer. What MARO does not show is a reliable correctness advantage, and simpler paths share its completeness gain while using less than a third of its prompt tokens.

The most favorable protocol and a fuller one support different conclusions about the same saved answers. Run 1 alone, with Qwen, trace-aware grading and answerable questions only, shows C ahead of F by 0.233 (1.544 against 1.311). With EXAONE, the same narrow protocol shows C behind F by 0.022 (1.300 against 1.322). Repeated, paired, all-question estimates range from \ensuremath{-}0.118 to +0.060, and no interval excludes zero.

\subsection{Which conclusions depend on the protocol}

Correctness rankings were the most fragile result. They changed with repetition, generator, judge packet and scoring population. Completeness was the most robust: it favored C under trace-aware grading, answer-only grading and blinded human ratings. The simplified paths A and B, however, gained a similar amount over F (Section 6.9). Cost was stable by construction, at no less than 3.4 times F's model calls.

This pattern separates coverage from correctness. Drafting candidates and synthesizing them, which A, B and C share, is associated with more complete answers than vanilla RAG's single grounded call; C's planner, selector and verifier add no measurable completeness beyond A and B (Section 6.9). That does not make its claims more accurate, and it does not show that the added calls are worth paying for.

\subsection{Why the judge packet matters}

Identical C and E answers received different scores when only the trace differed. Agent-as-a-Judge (Zhuge et al., 2025) treats access to intermediate steps as a strength, because it lets the judge check process. Our results add a caution: process information also moves the score of an unchanged answer. A judge that sees a multi-stage trace is evaluating the system's visible work as well as its output.

We cannot say from these data that trace-aware grading systematically favors multi-stage systems. The answer-only audit changed traces, source lists and the endpoint together, and run 1 moved most. What the data do show is that the principal estimate depends on the packet. Trace-aware results should therefore be accompanied by an answer-only check on the same answers.

\subsection{Hedging is a distinct behavior}

C's unanswerable responses were mostly hedges: they noted the missing evidence and then supplied a nearby figure. For a policy reader, a hedge that cites another year's value can mislead as much as a confident wrong answer. It is not the same failure as a false refusal, which withholds an answer the collection supports. F showed the opposite pattern, refusing about a fifth of answerable questions.

The guard in G and I turned some hedges into clean refusals while adding some false refusals. Whether that trade is worth making depends on the cost of each error in the application. A binary abstention metric cannot inform that choice, because it merges hedges with one of the other two categories.

\subsection{Graph context, expansion and verification are different interventions}

Adding graph-linked chunks to the candidate pool (H) did not yield a clear correctness gain over C; this applies to the implemented traversal and reranking, not to graph retrieval generally. A diagnostic verifier can expose problems without fixing them: C and E return identical answers, so the experiment measures the verifier's cost and reporting, not any correction of text. Only G and I connect verification to the output, and their refusal patterns show both the value and the cost of doing so.

The error analysis (Section 6.8) does not support subqueries that drop temporal or regional anchors as the cause of C's item-level losses: wrong-period and wrong-region errors were mostly F's, and C's decisive losses came from hedged, substituted or unsupported added content, including one invented detail traced to a candidate draft. The blinded GPT-5 re-code agrees on the refusals and substitutions but codes five of C's losses as wrong region or indicator, against one in the original coding, so this conclusion is firmer for F's losses than for C's. It does support the hedging and false-refusal accounts in Section 8.3. Explanations for LightRAG's correctness deficit remain hypotheses, because J was not analysed.

\subsection{Recommended evaluation protocol}

For comparing RAG architectures on a domain benchmark, we recommend the following:

\begin{enumerate}
\def\labelenumi{\arabic{enumi}.}
\tightlist
\item
  Generate each configuration at least three times, report per-run contrasts, and average within question before averaging across questions.
\item
  Pair comparisons at the question level, resample questions for intervals, and state which effect sizes each interval rules out.
\item
  Replicate the headline comparison with a second generator family.
\item
  When the judge packet includes traces or system-specific sources, re-grade the key comparison with an answer-only common packet. Where two configurations share answer strings, use the pair as a packet-sensitivity control.
\item
  Report answerable-only and all-question results separately.
\item
  Code responses as refusal, hedge or answer on both answerable and unanswerable questions, and report false refusals.
\item
  Compare retrieval at matched source-text budgets, with the uncapped ceiling alongside.
\item
  Report criteria separately, and validate each criterion against human ratings.
\item
  Report model calls and prompt and completion tokens for every configuration.
\end{enumerate}

\subsection{Implications for policy information systems}

The evaluated systems occupy different operating points, and the right one depends on the application.

\begin{longtable}[]{@{}
  >{\raggedright\arraybackslash}p{(\linewidth - 4\tabcolsep) * \real{0.3333}}
  >{\raggedright\arraybackslash}p{(\linewidth - 4\tabcolsep) * \real{0.3333}}
  >{\raggedright\arraybackslash}p{(\linewidth - 4\tabcolsep) * \real{0.3333}}@{}}
\caption{Application priorities, observations in this study and engineering implications.}\tabularnewline
\toprule\noalign{}
\begin{minipage}[b]{\linewidth}\raggedright
Application priority
\end{minipage} & \begin{minipage}[b]{\linewidth}\raggedright
Observation in this study
\end{minipage} & \begin{minipage}[b]{\linewidth}\raggedright
Implication
\end{minipage} \\
\midrule\noalign{}
\endfirsthead
\toprule\noalign{}
\begin{minipage}[b]{\linewidth}\raggedright
Application priority
\end{minipage} & \begin{minipage}[b]{\linewidth}\raggedright
Observation in this study
\end{minipage} & \begin{minipage}[b]{\linewidth}\raggedright
Implication
\end{minipage} \\
\midrule\noalign{}
\endhead
\bottomrule\noalign{}
\endlastfoot
Broad coverage and correct reporting periods & Completeness and human temporal accuracy favor C & Check whether these gains justify at least 3.4 times the model calls \\
Conservative handling of absent facts & F refuses all unanswerable items but also 18--19 of 90 answerable items per run & Measure false refusals alongside correct refusals \\
Intermediate refusal behavior & G and I refuse fewer answerable items than F and at least half of unanswerable items (G 5--7, I 8 of 10 per run) & Treat the guard as a policy choice with measurable costs \\
Traceable quality checks & C's verifier records judgments but never changes the answer & Distinguish monitoring from an intervention that changes output \\
Stable architecture decisions & Run-1 contrasts shrank or reversed later & Keep repeated outputs and report paired item-level uncertainty \\
\end{longtable}

These are candidates for deployment testing, not evidence of better policy decisions. A production evaluation should also measure task completion, source inspection, and the consequences of wrong answers and refusals.

\subsection{Limitations}

\textbf{Scope.} The study uses one Korean report corpus and 100 authored questions, with only ten unanswerable items. Category proportions are deliberate and may differ from real workloads. Gold evidence may omit other valid passages. Findings concern the evaluated system bundles: C and F differ in retrieval, prompting and orchestration, so no result isolates the effect of orchestration.

\textbf{Repetition.} Three runs per configuration give limited information about the generation distribution, and E is derived rather than generated. Repeated runs improve the estimate for these questions but do not increase the number of independent questions beyond 100.

\textbf{Generator replication.} The second generator replicates only the C/F comparison; the ablations, refusal coding and human study use Qwen alone. The two EXAONE deployments are quantized (8-bit MLX and FP8) on different servers, so they are not a controlled study of quantization. Seven EXAONE C grades on two unanswerable questions failed and remain ungraded.

\textbf{Judging.} A single judge model was used. The answer-only audit changed traces, source lists and the serving endpoint together, and the endpoint's equivalence to the original judge was not verified. It therefore tests robustness to a bundled protocol change, not trace influence alone, and it covers only Qwen C and F. Evidence support in the audit is judged against common gold passages, which may penalize valid support outside them. One Qwen judge packet used a recorded temperature deviation and one Qwen and one answer-only grade remain missing. Judge test-retest reliability on identical packets was not measured.

\textbf{Scoring and statistics.} Correctness uses a coarse three-point scale with many ties. The all-question endpoint mixes answer quality with refusal behavior. Character budgets were chosen post hoc, count characters rather than tokens, and credit only fully included passages; returned source text is a proxy for, not a copy of, the generation context. Intervals are nominal, unadjusted for multiple contrasts, and conditional on the saved runs and the reviewer panel.

\textbf{Retrieval comparability.} Matched character budgets improve comparison of returned evidence, but LightRAG's block expansion, aliases, ordering and prompt construction still differ. This is not an equal-prompt-budget generation experiment. The internal configurations and vanilla RAG share one embedder, BGE-M3; unlike Akarsu et al.~(2026), we did not vary it, so the retrieval and answer contrasts may depend on that choice.

\textbf{Human study.} Ratings cover run-1 C and F only. Reviewer and domain are partly confounded, readability coverage is incomplete, and human--human agreement is modest. One reviewer's unanswerable cards were re-rated after an instruction clarification, and an interface defect required re-entry of those scores; both are recorded.

\textbf{Refusal coding and cost.} Refusal behavior was coded by a fixed keyword rule. Human verdicts validate only its clean-refusal decision, for run-1 C and F (\ensuremath{\kappa} = 0.675); no human checked the hedge/answer boundary, and a refusal or hedge phrased outside the rule's word list would be miscoded. Resource figures exclude index construction, and call counts are not monetary cost. The Section 6.6 claim labels and Section 6.8 error categories were checked by blinded GPT-5 re-codes (\ensuremath{\kappa} = 0.86 and 0.63), not by a second human coder.

\section{Conclusion}

We presented MARO, a five-agent RAG system that answers Korean policy questions from 245 regional reports with cited answers that can be inspected stage by stage, and evaluated it against vanilla RAG, LightRAG and simplified variants of itself.

On our Korean policy-report benchmark, the verdict on MARO depended on the evaluation protocol. Its correctness advantage over vanilla RAG appeared under the narrowest protocol and disappeared under fuller ones: across repeated runs, a second generator family and answer-only grading, the paired contrast ranged from \ensuremath{-}0.118 to +0.060 and no interval excluded zero. Across all eleven paired contrasts in the study, no correctness verdict held under every protocol the data support. Identical answers received different scores depending on the trace the judge saw. Scoring population reversed the comparison of simplified paths with vanilla RAG. Three-way coding showed that the pipeline hedges where vanilla RAG refuses, and that vanilla RAG also refuses about a fifth of answerable questions.

Completeness was the most stable result: the pipeline gave more complete answers than vanilla RAG under every protocol, and blinded human ratings of one generation run agreed. The two simplified paths, without planning, selection or verification, gained a similar amount over vanilla RAG (Section 6.9), so in this study the pipeline's additional stages, at no less than 3.4 times the model calls, added no measurable completeness. Human reviewers also found its reporting periods more accurate and vanilla RAG's answers more readable.

The broader lesson is about measurement. Architecture claims for RAG should rest on repeated, paired, criterion-specific comparisons, checked against a second generator and an answer-only judge packet, with refusals, hedges and costs reported explicitly. A single correctness ranking from one run and one judge packet is not enough to choose an architecture. The benchmark and saved artifacts are released as described in the data-availability statement.

\section*{CRediT authorship contribution statement}
\textbf{Saksonita Khoeurn:} Conceptualization, Methodology, Software, Formal analysis, Writing -- original draft. \textbf{In Hwan Oh:} Data curation, Validation. \textbf{Sovandara Chhoun:} Software, Validation. \textbf{Phannet Pov:} Software. \textbf{Wan-Sup Cho:} Data curation, Supervision.

\section*{Declaration of competing interest}
The authors declare that they have no known competing financial interests or personal relationships that could have appeared to influence the work reported in this paper.

\section*{Funding}
This research did not receive any specific grant from funding agencies in the public, commercial, or not-for-profit sectors.

\section*{Acknowledgments}
The authors thank the Chungbuk Research Institute for its collaboration on this project. We also thank Sujin Yeon and Taeyun Kim for ensuring the quality of the parsed data by reviewing the parsed corpus.

\section*{Supplementary material}
\addcontentsline{toc}{section}{Supplementary material}

Supplementary Material S1 gives the verified settings, analysis rules, wall-clock timings (shared server, uncontrolled load) and remaining reproducibility limits. It is distributed with this preprint as the ancillary file \texttt{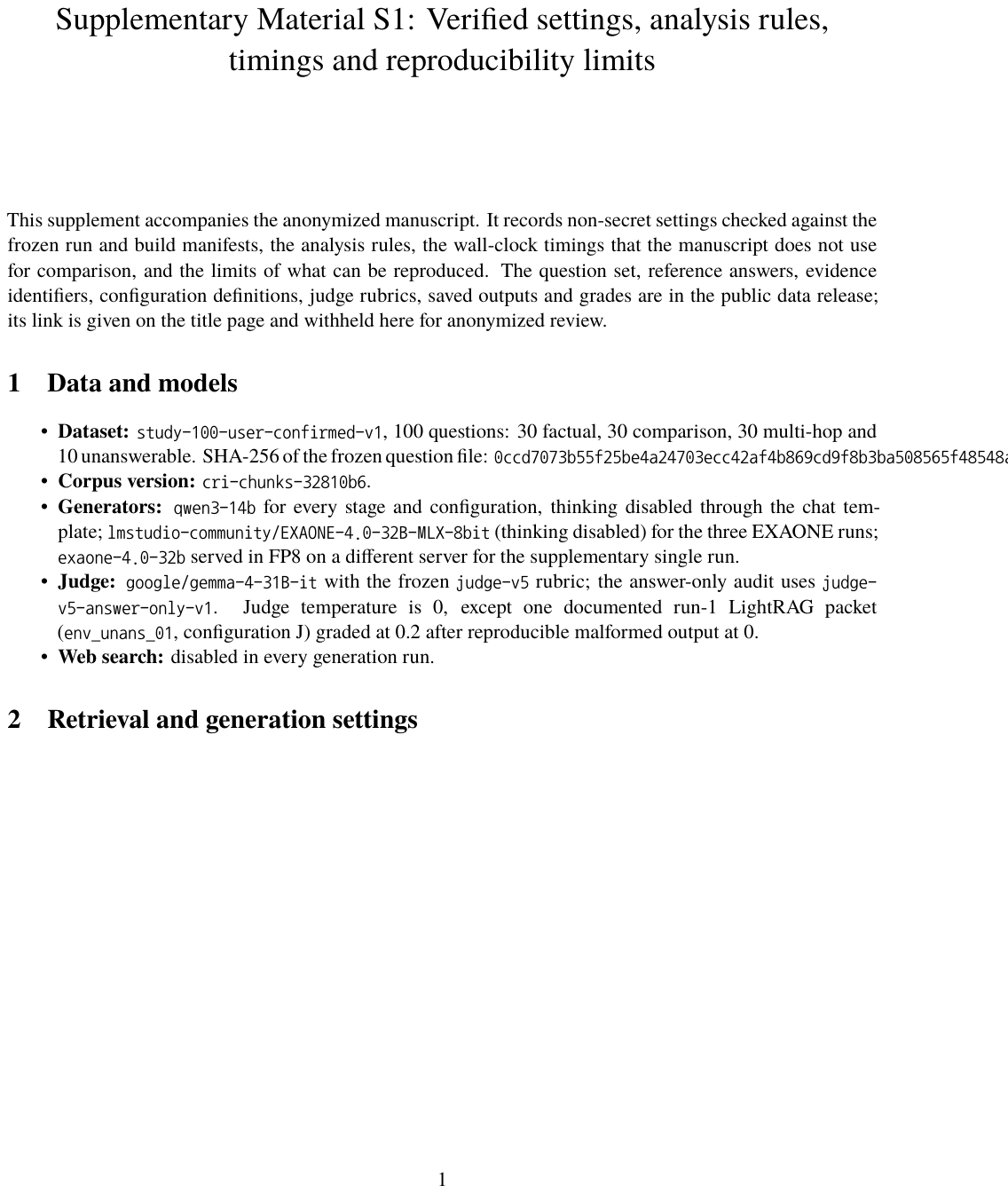}. It does not include unpublished question records, individual human ratings or answer artifacts.

\section*{Data availability}
\addcontentsline{toc}{section}{Data availability}

The benchmark and evaluation artifacts are publicly available at \url{https://github.com/cambigdatalabs/cri_dataset_for_paper}. The release includes the frozen question set, reference answers, evidence identifiers, dataset and corpus hashes, configuration definitions, judge rubrics (judge-v5 and judge-v5-answer-only-v1), saved outputs and grades, the error-analysis codings with the blinded re-code prompts and outputs, and reporting code. The source reports are third-party publications and are not redistributed: passages are identified by document, chunk and page, with a SHA-256 hash of each passage text. Credentials, model endpoints, reviewer authentication material, human ratings and model weights are excluded. 

\section*{Ethics statement}
\addcontentsline{toc}{section}{Ethics statement}

Six professional reviewers assessed system-generated answers; reviewer identifiers are pseudonymized. 

\section*{Declaration of generative AI and AI-assisted technologies in the manuscript preparation process}
\addcontentsline{toc}{section}{Declaration of generative AI and AI-assisted technologies in the manuscript preparation process}

During the preparation of this work the authors used OpenAI Codex and Claude (Anthropic) to reorganize drafted manuscript text, check numerical consistency against local evaluation artifacts, and locate relevant literature. After using these tools, the authors reviewed and edited the content and take full responsibility for the content of the publication. The language models used for answer generation and grading are described in Sections 4 and 5, and those used to assist and re-code the error analyses in Sections 6.6 and 6.8.

\section*{References}
\begingroup
\setlength{\parindent}{0pt}\setlength{\parskip}{3pt}\small\RaggedRight
\everypar{\hangindent=1.5em\hangafter=1}
Akarsu, M., \"Ozdemir, B., B\"uy\"uk\c{c}olak, D., \& Karaman, R. K. (2026). A triple-robustness analysis of retrieval-augmented generation for multi-hop requirements traceability {[}Preprint{]}. arXiv:2608.00705. \url{https://doi.org/10.48550/arXiv.2608.00705}

Allganize. (2024). RAG-Evaluation-Dataset-KO {[}Dataset{]}. Hugging Face. \url{https://huggingface.co/datasets/allganize/RAG-Evaluation-Dataset-KO}

Asai, A., Wu, Z., Wang, Y., Sil, A., \& Hajishirzi, H. (2024). Self-RAG: Learning to retrieve, generate, and critique through self-reflection. In The Twelfth International Conference on Learning Representations (ICLR 2024). \url{https://openreview.net/forum?id=hSyW5go0v8}

Atil, B., Aykent, S., Chittams, A., Fu, L., Passonneau, R. J., Radcliffe, E., Rajagopal, G. R., Sloan, A., Tudrej, T., Ture, F., Wu, Z., Xu, L., \& Baldwin, B. (2025). Non-determinism of ``deterministic'' LLM system settings in hosted environments. In Proceedings of the 5th Workshop on Evaluation and Comparison of NLP Systems (Eval4NLP 2025) (pp.~135--148). Association for Computational Linguistics. \url{https://aclanthology.org/2025.eval4nlp-1.12/}

Bae, K., Choi, E., Choi, K., Choi, S. J., Choi, Y., Han, K., et al.~(2025). EXAONE 4.0: Unified large language models integrating non-reasoning and reasoning modes {[}Preprint{]}. arXiv:2507.11407. \url{https://doi.org/10.48550/arXiv.2507.11407}

Bell, S. J., Kampman, O. P., Dodge, J., \& Lawrence, N. D. (2022). Modeling the machine learning multiverse. In Advances in Neural Information Processing Systems 35 (NeurIPS 2022). \url{https://doi.org/10.48550/arXiv.2206.05985}

Chang, C.-Y., Jiang, Z., Rakesh, V., Pan, M., Yeh, C.-C. M., Wang, G., Hu, M., Xu, Z., Zheng, Y., Das, M., \& Zou, N. (2025). MAIN-RAG: Multi-agent filtering retrieval-augmented generation. In Proceedings of the 63rd Annual Meeting of the Association for Computational Linguistics (Volume 1: Long Papers) (pp.~2607--2622). Association for Computational Linguistics. \url{https://doi.org/10.18653/v1/2025.acl-long.131}

Chen, J., Lin, H., Han, X., \& Sun, L. (2024). Benchmarking large language models in retrieval-augmented generation. Proceedings of the AAAI Conference on Artificial Intelligence, 38(16), 17754--17762. \url{https://doi.org/10.1609/aaai.v38i16.29728}

Chen, J., Xiao, S., Zhang, P., Luo, K., Lian, D., \& Liu, Z. (2024). M3-embedding: Multi-linguality, multi-functionality, multi-granularity text embeddings through self-knowledge distillation. In Findings of the Association for Computational Linguistics: ACL 2024 (pp.~2318--2335). Association for Computational Linguistics. \url{https://doi.org/10.18653/v1/2024.findings-acl.137}

Dimitrova, M. (2025). Multi-agent coordination strategies vs.~retrieval-augmented generation in LLMs: A comparative evaluation. Electronics, 14(24), 4883. \url{https://doi.org/10.3390/electronics14244883}

Edge, D., Trinh, H., Cheng, N., Bradley, J., Chao, A., Mody, A., Truitt, S., Metropolitansky, D., Ness, R. O., \& Larson, J. (2024). From local to global: A graph RAG approach to query-focused summarization {[}Preprint{]}. arXiv:2404.16130. \url{https://doi.org/10.48550/arXiv.2404.16130}

Elkoussy, L., \& Perez, J. (2026). AgentLTL: A trace-verification framework for measuring, enforcing, and training procedural compliance in tool-using LLM agents {[}Preprint{]}. arXiv:2607.02599. \url{https://doi.org/10.48550/arXiv.2607.02599}

Es, S., James, J., Espinosa Anke, L., \& Schockaert, S. (2024). RAGAS: Automated evaluation of retrieval augmented generation. In Proceedings of the 18th Conference of the European Chapter of the Association for Computational Linguistics: System Demonstrations (pp.~150--158). Association for Computational Linguistics. \url{https://doi.org/10.18653/v1/2024.eacl-demo.16}

Fan, D., Xue, Z., Liu, S., \& Tan, Q. (2026). Do we still need GraphRAG? Benchmarking RAG and GraphRAG for agentic search systems {[}Preprint{]}. arXiv:2604.09666. \url{https://doi.org/10.48550/arXiv.2604.09666}

Ferrazzi, P., Cvjeti\'canin, M., Piraccini, A., \& Giannuzzi, D. (2026). Is agentic RAG worth it? An experimental comparison of RAG approaches. In Proceedings of the 64th Annual Meeting of the Association for Computational Linguistics (Industry Track) (pp.~56--75). Association for Computational Linguistics. \url{https://doi.org/10.18653/v1/2026.acl-industry.5}

Gan, A., Yu, H., Zhang, K., Liu, Q., Yan, W., Huang, Z., Tong, S., \& Hu, G. (2025). Retrieval augmented generation evaluation in the era of large language models: A comprehensive survey {[}Preprint{]}. arXiv:2504.14891. \url{https://doi.org/10.48550/arXiv.2504.14891}

Gao, T., Yen, H., Yu, J., \& Chen, D. (2023). Enabling large language models to generate text with citations. In Proceedings of the 2023 Conference on Empirical Methods in Natural Language Processing (pp.~6465--6488). \url{https://doi.org/10.18653/v1/2023.emnlp-main.398}

Guo, Z., Xia, L., Yu, Y., Ao, T., \& Huang, C. (2025). LightRAG: Simple and fast retrieval-augmented generation. In Findings of the Association for Computational Linguistics: EMNLP 2025 (pp.~10746--10761). Association for Computational Linguistics. \url{https://doi.org/10.18653/v1/2025.findings-emnlp.568}

Guti\'errez, B. J., Shu, Y., Qi, W., Zhou, S., \& Su, Y. (2025). From RAG to memory: Non-parametric continual learning for large language models. In Proceedings of the 42nd International Conference on Machine Learning (Vol. 267, pp.~21497--21515). PMLR. \url{https://proceedings.mlr.press/v267/gutierrez25a.html}

Han, H., Ma, L., Wang, Y., Shomer, H., Lei, Y., Qi, Z., Guo, K., Hua, Z., Long, B., Liu, H., Aggarwal, C. C., \& Tang, J. (2025). RAG vs.~GraphRAG: A systematic evaluation and key insights {[}Preprint{]}. arXiv:2502.11371. \url{https://doi.org/10.48550/arXiv.2502.11371}

Heineman, D., Hofmann, V., Magnusson, I., Gu, Y., Smith, N. A., Hajishirzi, H., Lo, K., \& Dodge, J. (2025). Signal and noise: A framework for reducing uncertainty in language model evaluation {[}Preprint{]}. arXiv:2508.13144. \url{https://doi.org/10.48550/arXiv.2508.13144}

Hochlehnert, A., Bhatnagar, H., Udandarao, V., Albanie, S., Prabhu, A., \& Bethge, M. (2025). A sober look at progress in language model reasoning: Pitfalls and paths to reproducibility. In Conference on Language Modeling (COLM 2025). \url{https://doi.org/10.48550/arXiv.2504.07086}

Kapoor, S., Stroebl, B., Siegel, Z. S., Nadgir, N., \& Narayanan, A. (2025). AI agents that matter. Transactions on Machine Learning Research. \url{https://doi.org/10.48550/arXiv.2407.01502}

Kim, Y., Choi, Y. R., Choi, E., Choi, J., Park, H. J., \& Hwang, W. (2024). Developing a pragmatic benchmark for assessing Korean legal language understanding in large language models. In Findings of the Association for Computational Linguistics: EMNLP 2024. Association for Computational Linguistics. \url{https://doi.org/10.18653/v1/2024.findings-emnlp.319}

Kim, Y., Yeen, H., Yun, H., \& Lee, J. (2025). Ko-LongRAG: A Korean long-context RAG benchmark built with a retrieval-free approach. In Findings of the Association for Computational Linguistics: EMNLP 2025. Association for Computational Linguistics. \url{https://doi.org/10.18653/v1/2025.findings-emnlp.938}

Koco\'n, J., Piasecki, M., Janz, A., Ferdinan, T., Radli\'nski, \L{}., Koptyra, B., Oleksy, M., Wo\'zniak, S., Walkowiak, P., Wojtasik, K., Moska, J., Naskr\k{e}t, T., Walkowiak, B., Gniewkowski, M., Szyc, K., Motyka, D., Banach, D., Dalasi\'nski, J., Rudnicka, E., . . . P\k{e}zik, P. (2025). PLLuM: A family of Polish large language models {[}Preprint{]}. arXiv:2511.03823. \url{https://doi.org/10.48550/arXiv.2511.03823}

Lee, J., Kim, S., Park, W., Lee, G., Kim, S., \& Lee, M. (2025). SDS KoPub VDR: A benchmark dataset for visual document retrieval in Korean public documents {[}Preprint{]}. arXiv:2511.04910. \url{https://doi.org/10.48550/arXiv.2511.04910}

Lewis, P., Perez, E., Piktus, A., Petroni, F., Karpukhin, V., Goyal, N., K\"uttler, H., Lewis, M., Yih, W.-t., Rockt\"aschel, T., Riedel, S., \& Kiela, D. (2020). Retrieval-augmented generation for knowledge-intensive NLP tasks. In Advances in Neural Information Processing Systems 33 (pp.~9459--9474). Curran Associates. \url{https://doi.org/10.48550/arXiv.2005.11401}

Lu, W., Chen, K., Shen, Z., Qiao, R., \& Sun, X. (2026). HiChunk: Evaluating and enhancing retrieval augmented generation with hierarchical chunking. In Proceedings of the 64th Annual Meeting of the Association for Computational Linguistics (Volume 1: Long Papers) (pp.~29738--29753). Association for Computational Linguistics. \url{https://doi.org/10.18653/v1/2026.acl-long.1372}

Madaan, L., Singh, A. K., Schaeffer, R., Poulton, A., Koyejo, S., Stenetorp, P., Narang, S., \& Hupkes, D. (2024). Quantifying variance in evaluation benchmarks {[}Preprint{]}. arXiv:2406.10229. \url{https://doi.org/10.48550/arXiv.2406.10229}

Marioriyad, A., Ghahroodi, O., Asgari, E., Rohban, M. H., \& Soleymani Baghshah, M. (2026). The judge who never admits: Hidden shortcuts in LLM-based evaluation {[}Preprint{]}. arXiv:2602.07996. \url{https://doi.org/10.48550/arXiv.2602.07996}

Mathur, S., Rittner, R. D., Thakur, V. A., Schiff, D. S., \& Islam, T. (2026). Retrieval improvements do not guarantee better answers: A study of RAG for AI policy QA {[}Preprint{]}. arXiv:2603.24580. \url{https://doi.org/10.48550/arXiv.2603.24580}

Miller, E. (2024). Adding error bars to evals: A statistical approach to language model evaluations {[}Preprint{]}. arXiv:2411.00640. \url{https://doi.org/10.48550/arXiv.2411.00640}

Muhamed, A., Ribeiro, L. F. R., Dreyer, M., Smith, V., \& Diab, M. T. (2026). RefusalBench: Generative evaluation of selective refusal in grounded language models. In Proceedings of the 19th Conference of the European Chapter of the Association for Computational Linguistics (Volume 1: Long Papers) (pp.~6811--6856). Association for Computational Linguistics. \url{https://doi.org/10.18653/v1/2026.eacl-long.321}

Nguyen, T., Chin, P., \& Tai, Y.-W. (2025). MA-RAG: Multi-agent retrieval-augmented generation via collaborative chain-of-thought reasoning {[}Preprint{]}. arXiv:2505.20096. \url{https://doi.org/10.48550/arXiv.2505.20096}

Peng, X., Choubey, P. K., Xiong, C., \& Wu, C.-S. (2025). Unanswerability evaluation for retrieval augmented generation. In Proceedings of the 63rd Annual Meeting of the Association for Computational Linguistics (Volume 1: Long Papers) (pp.~8452--8472). Association for Computational Linguistics. \url{https://doi.org/10.18653/v1/2025.acl-long.415}

R\"ottger, P., Kirk, H. R., Vidgen, B., Attanasio, G., Bianchi, F., \& Hovy, D. (2024). XSTest: A test suite for identifying exaggerated safety behaviours in large language models. In Proceedings of the 2024 Conference of the North American Chapter of the Association for Computational Linguistics: Human Language Technologies (Volume 1: Long Papers) (pp.~5377--5400). Association for Computational Linguistics. \url{https://doi.org/10.18653/v1/2024.naacl-long.301}

Ru, D., Qiu, L., Hu, X., Zhang, T., Shi, P., Chang, S., Jiayang, C., Wang, C., Sun, S., Li, H., Zhang, Z., Wang, B., Jiang, J., He, T., Wang, Z., Liu, P., Zhang, Y., \& Zhang, Z. (2024). RAGChecker: A fine-grained framework for diagnosing retrieval-augmented generation. In Advances in Neural Information Processing Systems 37 (Datasets and Benchmarks Track) (pp.~21999--22027). Curran Associates. \url{https://doi.org/10.52202/079017-0692}

Saad-Falcon, J., Khattab, O., Potts, C., \& Zaharia, M. (2024). ARES: An automated evaluation framework for retrieval-augmented generation systems. In Proceedings of the 2024 Conference of the North American Chapter of the Association for Computational Linguistics: Human Language Technologies (Volume 1: Long Papers) (pp.~338--354). Association for Computational Linguistics. \url{https://doi.org/10.18653/v1/2024.naacl-long.20}

Simonsohn, U., Simmons, J. P., \& Nelson, L. D. (2020). Specification curve analysis. Nature Human Behaviour, 4(11), 1208--1214. \url{https://doi.org/10.1038/s41562-020-0912-z}

Singh, A., Ehtesham, A., Kumar, S., Khoei, T. T., \& Vasilakos, A. V. (2025). Agentic retrieval-augmented generation: A survey on agentic RAG {[}Preprint{]}. arXiv:2501.09136. \url{https://doi.org/10.48550/arXiv.2501.09136}

Steegen, S., Tuerlinckx, F., Gelman, A., \& Vanpaemel, W. (2016). Increasing transparency through a multiverse analysis. Perspectives on Psychological Science, 11(5), 702--712. \url{https://doi.org/10.1177/1745691616658637}

Tamba, H. (2026). Necessary but not sufficient: Temperature control and reproducibility in LLM-as-judge safety evaluations {[}Preprint{]}. arXiv:2606.26185. \url{https://doi.org/10.48550/arXiv.2606.26185}

Wen, B., Yao, J., Feng, S., Xu, C., Tsvetkov, Y., Howe, B., \& Wang, L. L. (2025). Know your limits: A survey of abstention in large language models. Transactions of the Association for Computational Linguistics, 13, 529--556. \url{https://doi.org/10.1162/tacl/_a/_00754}

Xiang, Z., Wu, C., Zhang, Q., Chen, S., Hong, Z., Huang, X., \& Su, J. (2026). When to use graphs in RAG: A comprehensive analysis for graph retrieval-augmented generation. In International Conference on Learning Representations (ICLR 2026). \url{https://openreview.net/forum?id=i9q9xDMjG7}

Xie, R., Gopinath, D., Qiu, D., Lin, D., Sun, H., Potdar, S., \& Dhingra, B. (2026). Over-searching in search-augmented large language models. In Proceedings of the 19th Conference of the European Chapter of the Association for Computational Linguistics (Volume 1: Long Papers) (pp.~7714--7739). Association for Computational Linguistics. \url{https://doi.org/10.18653/v1/2026.eacl-long.361}

Xu, T., Zheng, H., Li, C., Chen, H., Liu, Y., Chen, R., \& Sun, L. (2025). NodeRAG: Structuring graph-based RAG with heterogeneous nodes {[}Preprint{]}. arXiv:2504.11544. \url{https://doi.org/10.48550/arXiv.2504.11544}

Yan, S.-Q., Gu, J.-C., Zhu, Y., \& Ling, Z.-H. (2024). Corrective retrieval augmented generation {[}Preprint{]}. arXiv:2401.15884. \url{https://doi.org/10.48550/arXiv.2401.15884}

Yang, A., Li, A., Yang, B., Zhang, B., Hui, B., Zheng, B., et al.~(2025). Qwen3 technical report {[}Preprint{]}. arXiv:2505.09388. \url{https://doi.org/10.48550/arXiv.2505.09388}

Ye, J., Wang, Y., Huang, Y., Chen, D., Zhang, Q., Moniz, N., Gao, T., Geyer, W., Huang, C., Chen, P.-Y., Chawla, N. V., \& Zhang, X. (2025). Justice or prejudice? Quantifying biases in LLM-as-a-judge. In International Conference on Learning Representations (ICLR 2025). \url{https://doi.org/10.48550/arXiv.2410.02736}

Zeng, Q., Yan, X., Luo, H., Lin, Y., Wang, Y., Fu, F., Du, B., Xu, Q., \& Jiang, J. (2025). How significant are the real performance gains? An unbiased evaluation framework for GraphRAG {[}Preprint{]}. arXiv:2506.06331. \url{https://doi.org/10.48550/arXiv.2506.06331}

Zhang, M., \& Choi, E. (2021). SituatedQA: Incorporating extra-linguistic contexts into QA. In Proceedings of the 2021 Conference on Empirical Methods in Natural Language Processing (pp.~7371--7387). \url{https://doi.org/10.18653/v1/2021.emnlp-main.586}

Zheng, L., Chiang, W.-L., Sheng, Y., Zhuang, S., Wu, Z., Zhuang, Y., Lin, Z., Li, Z., Li, D., Xing, E. P., Zhang, H., Gonzalez, J. E., \& Stoica, I. (2023). Judging LLM-as-a-judge with MT-Bench and Chatbot Arena. In Advances in Neural Information Processing Systems 36 (Datasets and Benchmarks Track). \url{https://proceedings.neurips.cc/paper/_files/paper/2023/hash/91f18a1287b398d378ef22505bf41832-Abstract.html}

Zhuge, M., Zhao, C., Ashley, D. R., Wang, W., Khizbullin, D., Xiong, Y., Liu, Z., Chang, E., Krishnamoorthi, R., Tian, Y., Shi, Y., Chandra, V., \& Schmidhuber, J. (2025). Agent-as-a-judge: Evaluate agents with agents. In Proceedings of the 42nd International Conference on Machine Learning (PMLR 267, pp.~80569--80611). \url{https://proceedings.mlr.press/v267/zhuge25a.html}

\endgroup

\end{document}